\documentclass{article}

\pdfoutput=1

    \usepackage[preprint]{neurips_2023}

\usepackage[utf8]{inputenc} 
\usepackage[T1]{fontenc}    
\usepackage{url}            
\usepackage{booktabs}       
\usepackage{amsfonts}       
\usepackage{nicefrac}       
\usepackage{microtype}      
\usepackage{xcolor}         

\usepackage{times}
\usepackage{epsfig}
\usepackage{graphicx}
\usepackage{amsmath}
\usepackage{amssymb}

\usepackage{subcaption}

\usepackage{booktabs}
\usepackage{stfloats}
\usepackage[normalem]{ulem}

\usepackage{footnotebackref}
\usepackage[para]{footmisc}
\usepackage{color}
\usepackage{enumerate}
\usepackage{graphbox}
\usepackage{makecell}
\usepackage{soul}
\usepackage{wrapfig}
\usepackage{graphics}
\usepackage{multirow}
\usepackage{eqparbox}
\usepackage{enumerate}

\usepackage{isotope}
\usepackage{siunitx}
\usepackage{tablists}
\usepackage{bm}
\usepackage{colortbl}  
\usepackage{array}   
\usepackage{enumitem}

\usepackage{booktabs}

\title{UniDiff: Advancing Vision-Language Models with Generative and Discriminative Learning}

\author{
 \textbf{Xiao~Dong}$^{1,2}$, \textbf{Runhui~Huang}$^1$, \textbf{Xiaoyong~Wei}$^2$, \textbf{Zequn~Jie}$^3$, \\ \textbf{Jianxing~Yu}$^1$, \textbf{Jian~Yin}$^1$, \textbf{Xiaodan~Liang}$^{1,2}$ \\
$^1$Sun Yat-Sen University,
$^2$Peng Cheng Laboratory,
$^3$Meituan}

\begin{document}

\maketitle

\begin{abstract}
Recent advances in vision-language pre-training have enabled machines to perform better in multimodal object discrimination (e.g., image-text semantic alignment) and image synthesis (e.g., text-to-image generation).  
On the other hand, fine-tuning pre-trained models with discriminative or generative capabilities such as CLIP and Stable Diffusion on domain-specific datasets has shown to be effective in various tasks by adapting to specific domains.  
However, few studies have explored the possibility of learning both discriminative and generative capabilities and leveraging their synergistic effects to create a powerful and personalized multimodal model during fine-tuning. 
This paper presents \textbf{UniDiff}, a unified multi-modal model that integrates image-text contrastive learning (ITC), text-conditioned image synthesis learning (IS), and reciprocal semantic consistency modeling (RSC).  
UniDiff effectively learns aligned semantics and mitigates the issue of semantic collapse during fine-tuning on small datasets by leveraging RSC on visual features from CLIP and diffusion models, without altering the pre-trained model's basic architecture.  
UniDiff demonstrates versatility in both multi-modal understanding and generative tasks.  
Experimental results on three datasets (Fashion-man, Fashion-woman, and E-commercial Product) showcase substantial enhancements in vision-language retrieval and text-to-image generation, illustrating the advantages of combining discriminative and generative fine-tuning.  
The proposed UniDiff model establishes a robust pipeline for personalized modeling and serves as a benchmark for future comparisons in the field. 
\end{abstract}

\vspace{-3mm}
\section{Introduction}
\label{sec:intro}


Recent advancements in computer vision have been driven by scaling models on large datasets of captioned images collected from the internet. CLIP \cite{radford2021learning} has emerged as a successful representation learner for aligning the semantics between image and text through contrastive learning and prompt modeling, which exhibits remarkable discriminative capabilities and has been fine-tuned to achieve state-of-the-art performance on various cross-modal tasks, including zero-shot classification and image-text retrieval.  
Concurrently, diffusion models~\cite{ddpm,adm,rombach2022high} have demonstrated promising performance through iterative refining generated samples, enhancing sample quality and providing greater control over the generation process. 
Furthermore, fine-tuning techniques~\cite{hulora,zaken2021bitfit} have shown significant performance on specific domain data beneficial of large-scale pre-trained discriminative or generative models instead of  training from scratch.  
In this work, we aim to integrate pre-trained discriminative and generative models seamlessly to align images and text based on their semantic meaning, while generating images guided by textual descriptions, resulting in a robust and personalized multi-modal model during fine-tuning process. 

Revisiting the research on unified discriminative and generative models, one famous and typical unified discriminative and generative model is generative adversarial networks (GAN) \cite{goodfellow2020generative}, which uses a discriminator to guide the adversarial process of the generator for better natural image synthesis.
However, this training mode often leads to model collapse and training instabilities.  
In contrast, diffusion-based generative models have demonstrated their superiority in unconditioned and multi-conditioned image generation tasks such as text-to-image, inpainting, and style transfer~\cite{rombach2022high}.   
Several works, such as DiffusionCLIP \cite{kim2022diffusionclip} and DiffuseIT \cite{kwon2022diffusion}, attempt to combine the CLIP model to achieve zero-shot image manipulation guided by text prompts.   
However, current CLIP-combined models solely focus on preserving the semantics of individual image-text pairs, overlooking the auxiliary information from other image-text pairs, and failing to achieve the discriminative task such as cross-modal retrieval and zero-shot classification.   
To fill the gap, a simple and efficient solution is to use a pre-trained CLIP model to guide the training of the diffusion model in an end-to-end manner, but direct employment of the CLIP model through image-text contrastive (ITC) learning may cause problems in semantic alignment learning due to a smaller training dataset (semantic collapse), as demonstrated in our ablation studies.  
 Therefore, exploring a simple and effective method to achieve a unified generative and discriminative model is meaningful and challenging.  

\begin{figure*}[t!]
\begin{center}
\vspace{-6mm}
\includegraphics[width=\linewidth]{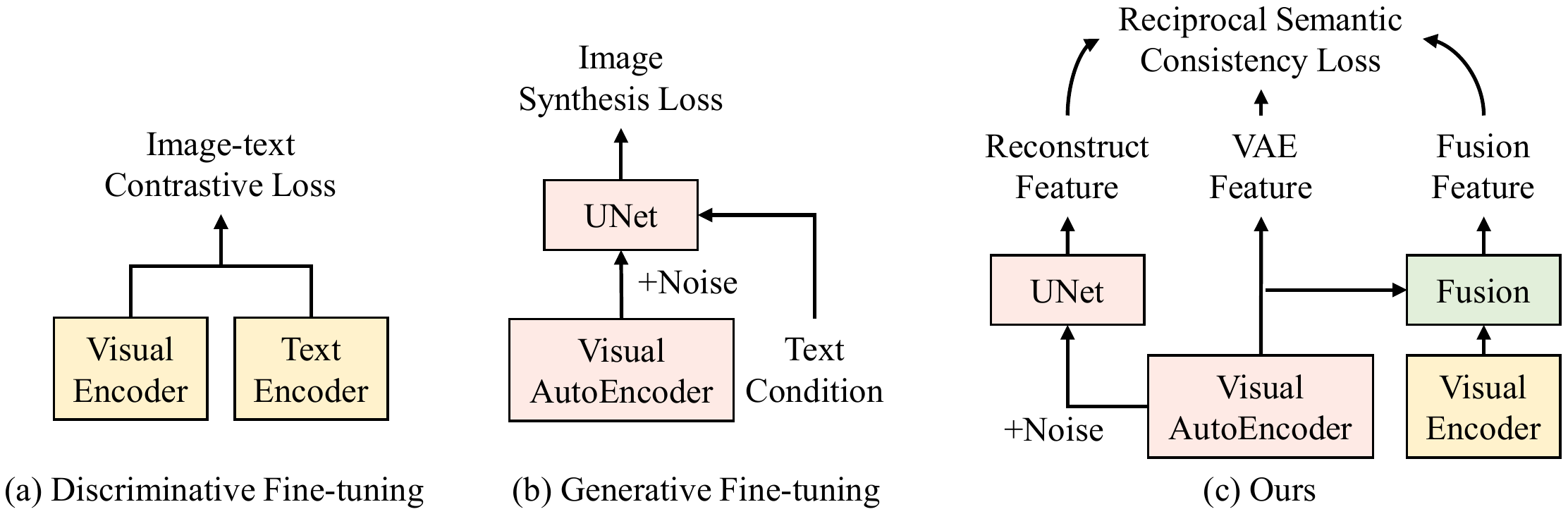}
\vspace{-6mm}
\end{center}
  \caption{Comparison of current fine-tuning methods and our method. (a) The discriminative fine-tuning method, like CLIP~\cite{radford2021learning}. (b) The generative fine-tuning method, like Stable Diffusion~\cite{rombach2022high}. (c) Our proposed Reciprocal Semantic Consistency Loss maintains the semantic consistency between discriminative visual features and generative visual features. Our method performs (a) discriminative fine-tuning, (b) generative fine-tuning and (c) reciprocal semantic consistency learning simultaneously.}
\label{fig:demo}
\vspace{-6mm}
\end{figure*}

In this paper, we introduce UniDiff, a novel unified framework that combines generative and discriminative approaches to align images and text.  
Figure~\ref{fig:demo} highlights the core concept of UniDiff and illustrates the distinctions between discriminative and generative models.  
UniDiff employs three key modeling techniques: image-text contrastive learning (ITC), text-conditioned image synthesis learning (IS), and reciprocal semantic consistency (RSC) modeling.  
 The primary objective of UniDiff is to achieve effective image-text alignment during fine-tuning while enhancing the overall quality of generated images.  
To overcome the challenge of semantic collapse during fine-tuning of ITC with limited training data, we introduce RSC modeling in UniDiff, inspired by style transfer learning.  
RSC enhances the semantic alignment of visual features extracted from both the CLIP and diffusion models, effectively bridging the semantic gap between these two visual encoders.  
By aligning reconstructed image features with text input, fused features, and semantic-level features learned from an AutoEncoder, RSC ensures a powerful and stable semantic understanding capability. 
UniDiff also introduces text-conditional generative learning, which integrates token-level semantics into the diffusion model, enabling the generation of high-fidelity images closely aligned with the provided text.  
Furthermore, UniDiff incorporates a lightweight module that plays a vital role in reconciling and integrating the visual features from both CLIP's visual encoder and Diffusion's autoencoder, thus bridging the feature gap between them. 
By incorporating RSC, UniDiff addresses the challenge of semantic collapse during fine-tuning on small datasets and improves both the quality of generative images and the alignment of image-text semantics.

Extensive experiments have confirmed the effectiveness of the proposed UniDiff method compared to the Stable Diffusion and surpassing other fine-tuning methods, such as LoRA~\cite{hulora}, BitFit~\cite{zaken2021bitfit}, and AdaptFormer~\cite{chen2022adaptformer}, on three special domain datasets.  
UniDiff demonstrates significant advancements in cross-modal retrieval and text-to-image generation even when trained with a small batch size, surpassing discriminative and generative methods designed for single-task training.   
Our work presents a pioneering technique that integrates the tasks of `understanding‘ and `painting’ within a comprehensive framework.  
This unified approach enables users to train personalized models without requiring extensive modifications to the architectures of pre-trained models. In summary, our contributions encompass three main aspects: 
\begin{itemize}[left=0em,itemsep=2pt,topsep=0pt,parsep=0pt]

\item We propose UniDiff, a unified vision-language model that addresses both multi-modality generation and discrimination tasks. UniDiff leverages the power of diffusion models and incorporates vision and language modalities to achieve comprehensive understanding and generation capabilities.

\item We propose reciprocal semantic consistency modeling, which serves for maintaining the consistency of discriminative semantics and generative semantics when unifying discriminative and generative vision-language training on personalized datasets.

\item We train our UniDiff on three specialized domain datasets. Experimental results highlight the superiority of combining vision-language discriminative and generative training within the UniDiff framework, showcasing its potential for advancing vision and language tasks. 
\end{itemize}
\section{Related Work}
\label{sec:related_work}

\noindent \textbf{Vision-Language Pre-training Model.} 
Building upon the success of the pre-training-and-fine-tuning paradigm on language model pre-training~\cite{radford2018improving,devlin2019bert,liu2019roberta,radford2019language}, numerous studies have explored vision-language pre-training using large-scale web-crawled image-text pairs. 
Through pre-training on extensive image-text data, both CLIP~\cite{radford2021learning} and ALIGN~\cite{jia2021scaling} have exhibited promising performance in various semantic understanding tasks, including zero-shot classification and image-text retrieval. 
VLP models can be broadly categorized into three types based on their approaches to semantic alignment. 
Firstly, there are image-text contrastive learning models such as CLIP \cite{radford2021learning}, ALIGN \cite{jia2021scaling}, FILIP \cite{yao2022filip}, DeCLIP~\cite{li2021supervision}, PyramidCLIP~\cite{gao2022pyramidclip}, which achieve cross-modal alignment through vision-language contrastive learning. 
Secondly, Mask Modeling models like ViLBERT~\cite{lu2019vilbert}, LXMERT~\cite{tan2019lxmert}, UNITER~\cite{chen2020uniter}, M6~\cite{lin2021m6}, DALL-E~\cite{ramesh2021zero},  BEiT v3~\cite{wang2022beitv3} employ multi-modal mask modeling to recover masked image and text tokens independently.  Lastly, hybrid Modeling approaches like ALBEF~\cite{li2021albef}, BLIP~\cite{li2022blip}, CAPTURE\cite{zhan2021product1m}, and SCALE~\cite{dong2022m5product} simultaneously utilize multi-modal contrastive learning and mask modeling to achieve effective multi-modal pre-training. 
In this paper, we aim to enhance the discriminative capability of the CLIP model and the generative compatibility of Stable Diffusion by enabling bidirectional consistency between discriminative features and generative features. 



\noindent \textbf{Diffusion-based Text-to-Image Generation Model.} 
Recently, diffusion models have emerged as a compelling approach for generating high-quality images. These models encompass both a forward process (signal to noise) and a reverse process (noise to signal) and have showcased remarkable success in image synthesis, particularly with Denoising Diffusion Probabilistic Models (DDPM)~\cite{ddpm} and score-based generative models. 
Ablated Diffusion Model (ADM)~\cite{adm} has exhibited superior image synthesis quality compared to alternative methods, including Variational Autoencoders (VAEs)~\cite{maaloe2019biva, razavi2019generating}, flow-based models~\cite{ho2019flow++,chen2020vflow}, auto-regressive models~\cite{chen18PixelSNAIL,parmar2018image}, and GANs~\cite{karras2020analyzing, biggan}. 
The generative power of diffusion models stems from their inherent alignment with the inductive biases of image-like data, achieved through the implementation of a U-Net neural architecture. 
Beneficial of pre-trained CLIP~\cite{radford2021learning} designed explicitly for image-text alignment, diffusion models utilize CLIP's text encoder to enable text-guided image synthesis~\cite{nichol2021glide,rombach2022high}. Stable Diffusion~\cite{rombach2022high} is proposed to perform the diffusion process on latent image space instead of pixel image space which improves the training efficiency and saves computational resources. 
Recent studies \cite{hulora,zaken2021bitfit,adapt1,adapt2,adapt3} propose some fine-tuning methods to transfer Stable Diffusion from the general domain to the specific domain and improve the generation performance.  
Fine-tuning uses two main techniques: adaptor and partial parameter tuning. 
LoRA~\cite{hulora} is an adaptor method that has received attention in image generation by introducing learnable projections between transformer layers.  
Compared to the adapter methods, partial parameter tuning methods, like BitFit~\cite{zaken2021bitfit}, only tune the bias parameters of each linear projection.  
 However, when dealing with training data that contains diverse categories, the limited number of trainable parameters may not effectively capture enough informative semantics. 
In our paper, we propose Reciprocal Semantic Consistency Learning to maintain semantic consistency between discriminative models and generative models, mutually enhancing each other's performance.
\section{Preliminary}
\label{sec:pre}
\subsection{Denoising Diffusion Probabilistic Models}
The DDPM (Denoising Diffusion Probabilistic Model)~\cite{ddpm} is a probabilistic model that consists of two Markov chains: the forward process and the backward process. 

\textbf{Forward Process.} The process involves generating a sequence of data by gradually adding Gaussian noise.  Starting with a sample $x_0$ drawn from a real-world data distribution $q(x_0)$, Gaussian noise $\epsilon \sim \mathcal{N}\left(0, \mathbf{I} \right)$ is incrementally added to $x_0$ at each step. 
The process can be formulated as follows:

\begin{equation}
q\left(\boldsymbol{x}_t \mid \boldsymbol{x}_{t-1}\right)=\mathcal{N}\left(\boldsymbol{x}_t ; \sqrt{1-\beta_t} \boldsymbol{x}_{t-1}, \beta_t \mathbf{I}\right),
\end{equation}
\begin{equation}
q\left(\boldsymbol{x}_{1: T} \mid \boldsymbol{x}_0\right)=\prod^T q\left(\boldsymbol{x}_t \mid \boldsymbol{x}_{t-1}\right),
\end{equation}
where $\beta \in \left(0, 1 \right)$ adjust the scale of the variance. 
$T$ is the maximum number of timesteps.  
According to the rule of sum normalized distributed random variables, we can sample any $x_t$ from $x_0$ with $q \left(x_t | x_0 \right) = \mathcal{N}\left(x_t, \sqrt{\bar{\alpha_t}}, (1-\bar{\alpha}_t, \left(1-\bar{\alpha}_t\mathbf{I} \right) \right)$, where $\alpha_t=1-\beta_t$ and $\bar{\alpha}=\prod_{s=1}^t \alpha_s$. 

\textbf{Reverse Process} 
The reverse process in DDPM starts from a Gaussian noise $X_T \sim \mathcal{N}\left(0, \mathbf{I} \right)$  and aims to learn a parameterized distribution $p_{\theta}\left(x_{t-1}| x_t \right)$ that can invert the diffusion process.   
The goal of the reverse process is to denoise an arbitrary Gaussian noise and recover a clean data sample. 
Mathematically, the reverse process can be described as follows:


\begin{equation}
    p_\theta\left(x_{t-1} \mid x_t\right)=\mathcal{N}\left(x_{t-1} ; \mu_\theta\left(x_t, t\right), \Sigma_\theta\left(x_t, t\right)\right), 
\end{equation}
\begin{equation}
    p_\theta\left(x_{0: T}\right)=p\left(x_T\right) \prod_{t=1}^T p_\theta\left(x_{t-1} \mid x_t\right), 
\end{equation}
where $p\left(x_T \right) = \mathcal{N}\left(x_T; 0, \mathbf{I} \right)$ and $\mu_\theta\left(x_t, t\right):=\frac{1}{\sqrt{\alpha_t}}\left(x_t-\frac{1-\alpha_t}{\sqrt{1-\bar{\alpha}_t}} \epsilon_\theta\left(x_t, t\right)\right)$.  

In the reverse process, we can approximately generate $x_0$ using one step with $\hat{x}_0=\left(x_t-\sqrt{1-\bar{\alpha}_t} \epsilon_\theta\left(x_t\right)\right) / \sqrt{\bar{\alpha}_t}$. 
 Here, $\epsilon_{\theta}$ is a function approximator that predicts $\epsilon$ from $x_t$.  
 The use of a Gaussian distribution $p_\theta\left(x_{t-1} \mid x_t\right)$ in the reverse process is justified by the fact that the reversal of the diffusion process has the same functional form as the forward process when $\beta_t$ (related to the diffusion step size) is small. 
 The generative distribution can be represented as $p_\theta\left(x_0\right)=\int p_\theta\left(x_{0: T}\right) d x_{1: T}$.

\textbf{Training.} During the training process, the objective is to maximize the log-likelihood of the model, which is achieved by minimizing the variational upper bound of the negative log-likelihood.  
This can be expressed as the integral of the product of the data distribution $q(\boldsymbol{x}_0)$ and the logarithm of the model distribution $p_\theta(\boldsymbol{x}_0)$ with respect to the variable $\boldsymbol{x}_0$. 
Following DDPM, the final objective is derived by some parameterization and simplification:
\begin{equation}
\label{eq:is_gene}
    \mathcal{L}_{\text {simple}}(\theta)=\mathbb{E}_{x_0, t, \epsilon}\left[\left\|\epsilon-\epsilon_\theta\right\|^2\right].
\end{equation}

\section{Method}
\label{sec:method}
\begin{figure*}[t!]
\begin{center}
\vspace{-2mm}
\includegraphics[width=\linewidth]{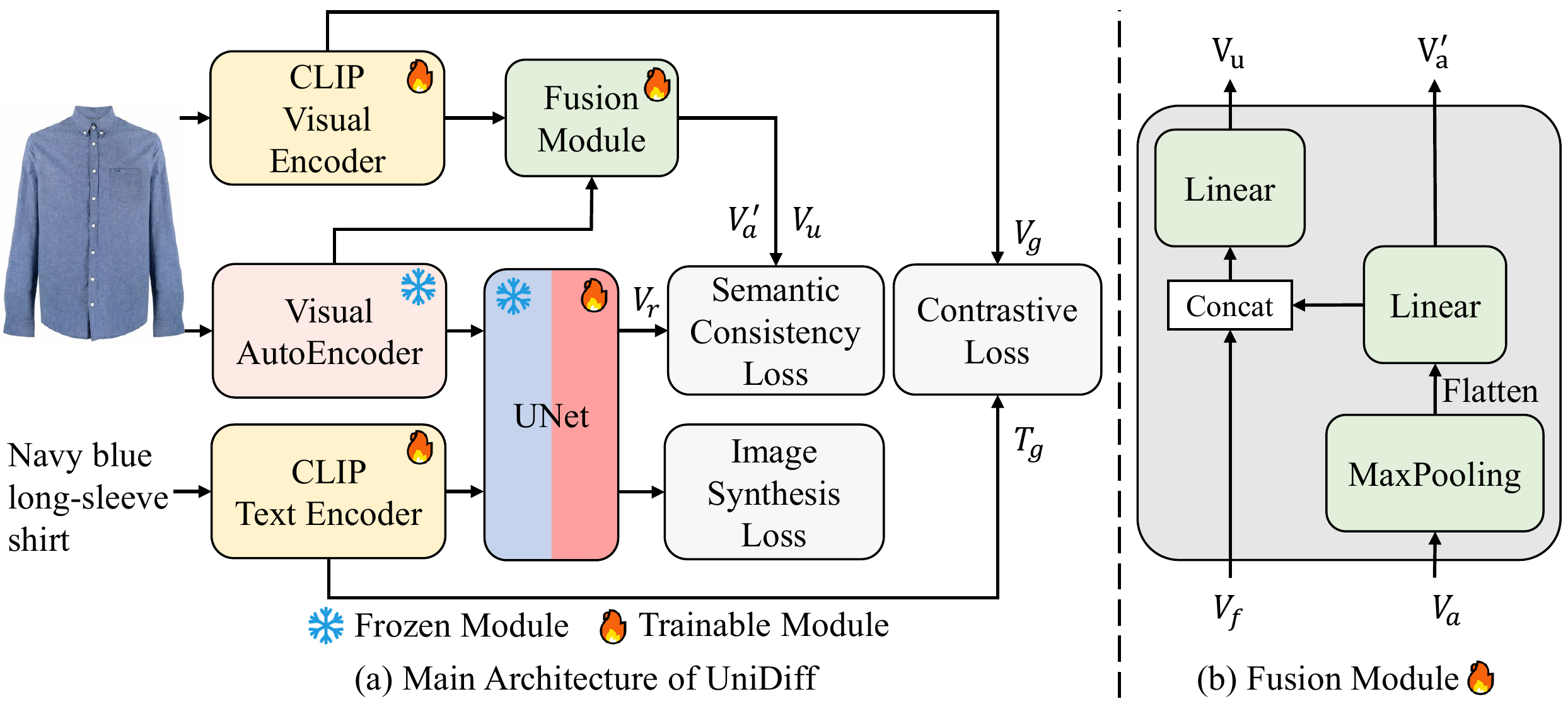}
\vspace{-6mm}
\end{center}
  \caption{Overall model architecture of UniDiff. Note that only the cross-attention module of the UNet is trainable, the other modules are frozen.}
\label{fig:overall_architecture}
\vspace{-7mm}
\end{figure*}

Given the remarkable performance of vision-language pre-trained models in zero-shot retrieval and classification tasks, we have chosen to utilize the widely used CLIP model as a guide for training our stable diffusion model.  
To achieve this, we incorporate a CLIP visual encoder and a designed fusion module into our generative model, resulting in CLIP-enhanced semantic guidance that significantly enhances the model's understanding capabilities.  
This enhancement has shown promising results across various downstream tasks. Our model, as depicted in Fig.~\ref{fig:overall_architecture}, exhibits image-text alignment, which is made possible by the integration of the CLIP model.  
In contrast to previous text-to-image generation models, our approach benefits from the advancements in vision-language alignment models and is trained to excel in both discriminative and generative tasks. 

\vspace{-3mm}
\subsection{Model Architecture}
\label{subsec:model_archi}

\textbf{Visual Encoder.} Given an input image $x$, we employ the visual encoder of CLIP for learning image-text alignment and the visual autoencoder of Stable Diffusion (SD)~\cite{rombach2022high} to perform the diffusion process on latent image space.  
The CLIP's visual encoder is trainable, while the parameters of the visual autoencoder remain frozen.   
These encoders enable us to generate two different types of features.  
Specifically, for the visual encoder of CLIP, we obtain token-level features $V_f \in \mathbb{R}^{l \times d }$ and the global feature $V_g \in \mathbb{R}^{d}$ of the image.  
Here, $l$ represents the number of image patches, and $d$ denotes the dimension of each token feature.  
The global feature corresponds to the $\mathbf{[CLS]}$ token feature. 
Additionally, we generate the normalized $\mathbf{[CLS]}$ feature to learn image-text alignment.  
We also use the autoencoder to obtain a three-dimensional representation of the features $V_a \in \mathbb{R}^{c \times h \times w}$, where $c, h, w$ represent the features' channel, height, and width, respectively.   
It is worth noting that the token-level feature $V_f$ in our approach includes the $\mathbf{[CLS]}$ token feature. 

\textbf{Text Encoder.} Given an input text $y$ paired with an image $x$, our initial step involves tokenizing it using the CLIP-tokenizer, resulting in a token sequence denoted as $W=\{w_i\}_{i=0}^m$. $m$ denotes the number of the word tokens. 
To obtain text features $T$, both for the entire text sentence and each token $w_i$, we utilize the CLIP's text encoder from the Stable Diffusion model.  
Notably, we allow the CLIP's text encoder to be trainable during the optimization process. 
Similar to image processing, we generate two types of outputs: 1) token-level features $T_f$ and 2) sentence-level features $T_g$ which is the $\mathbf{[EOS]}$ token feature. 
These features all undergo a post-processing step involving $\mathit{layernorm}$ and subsequent L2-normalization. 

\textbf{Fusion Module.} As illustrated in 
Fig.~\ref{fig:overall_architecture} (b), our fusion module takes two distinct feature inputs: $V_f$ and $V_a$, which are obtained from the CLIP's visual encoder and SD's visual autoencoder, respectively.  
 These features exhibit unique characteristics due to the different training tasks (discrimination and generation) of the original pre-trained model, resulting in variations in feature representations and semantic understanding.    
Specifically, $V_f$ primarily focuses on capturing the overall meaning of the sentence, while also providing a coarse token-level understanding.  
In contrast, $V_a$ represents three-dimensional features that excel in capturing detailed information for each token.  

To leverage the strengths of both features from distinct encoders, we apply a \textit{MaxPooling} operation with a stride of 2 and kernel size of 2 to enhance the edge feature.  
Subsequently, we consider the feature as a combination of $c$ maps $\in \mathbb{R}^{h/2 \times w/2}$ and flatten each map.   
To ensure compatibility with $V_f$, we use a linear layer to reduce the dimension from $(h/2) \times (w/2)$ to $d$.  
After passing through the linear layer, the features are concatenated along the first dimension.  
This concatenated feature in the space of $\mathbb{R}^{\left(l+c\right) \times d}$ is then transformed into a fused semantic feature $V_u \in \mathbb{R}^{n_u \times d}$ by a new linear layer that operates along the first dimension,  
where $n_u$ is the channel size of $V_u$.

\textbf{UNet.} Given the visual feature of the autoencoder $V_a$ and the token-level textual feature $T_f$, we employ the UNet module derived from the SD model.  
In the UNet, only the parameters of the cross-attention modules are trainable, while the parameters of the other modules are fixed. 
The UNet aims to denoise the noised $V_{a,t}$ with $t$-step noise under the textual condition of $T_f$. 
These features $V_{a,t}$ and $T_f$ are then fed into the UNet to predict the noise $\hat{\epsilon}=\epsilon_{\theta}(V_{a,t}, t, T_f)$ and  generate the $t$-step noised latent image $V_{a,t}$.  These outputs are subsequently used for network optimization.

\subsection{Optimization Function}
\label{subsec:opti_func}

Our main objective is to improve the accuracy and effectiveness of image-text alignment for discriminative tasks by incorporating token-level and text-level information.    
Furthermore, we employ aligned image-text representations to guide the generative task of the Stable Diffusion model, aiming to generate semantically matched and realistic images. 
To address the issue of semantic collapse caused by limited data in image-text contrastive (ITC) learning and to enhance the overall capabilities of the model, we introduce reciprocal semantic consistency (RSC) modeling.   
This approach bridges the gap between the visual features extracted from CLIP's visual encoder and the autoencoder of SD, effectively addressing the semantic collapse problem.  
In our UniDiff approach, we integrate the characteristics mentioned above to enable the vision-language model to develop a comprehensive understanding of both text and image through the integration of discriminative and generative tasks.

During the model optimization process, we employ a combination of three loss functions: an image-text contrastive loss $\mathcal{L}_{itc}$, a text-conditioned image synthesis loss $\mathcal{L}_{is}$, and a reciprocal semantic consistency loss $\mathcal{L}_{rsc}$. 
These loss functions collectively contribute to the refinement and alignment of the model's visual and textual representations. 

\textbf{Image-text Contrastive Loss.} 
For the loss $\mathcal{L}_{itc}$, we calculate the similarity between the image features $V_{g_i}$ and the text query features $T_{g_i}$ for a batch of $n$ image-text pairs.  
This computation generates a $n \times n$ similarity score map. 
Using this similarity score map, we then compute the image-to-text and text-to-image InfoNCE~\cite{oord2018representation} losses with learnable temperature $\tau_c$: 
\begin{equation}
\mathcal{L}_{i 2 t}=-\frac{1}{n} \sum_{i=1}^n \log \frac{\exp \left(V^{\top}_{g_i}{T_{g_i}} / \tau_c \right)}{\sum_j \exp \left(V^{\top}_{g_i}{T_{g_{j}}} / \tau_c \right)},
\end{equation}

\begin{equation}
\mathcal{L}_{t 2 i}=-\frac{1}{n} \sum_{j=1}^n \log \frac{\exp \left(V^{\top}_{g_j}{T_{g_j}}/ \tau_c \right)}
{\sum_i \exp \left(V^{\top}_{g_i} T_{g_j} / \tau_c \right)}.
\end{equation}
We then combine these losses into our symmetric image-text contrastive loss $\mathcal{L}_{\mathrm{itc}}$:
\begin{equation}
\label{eq:ITC}
\mathcal{L}_{\mathrm{itc}}=\frac{1}{2}\left(\mathcal{L}_{i 2 t}+\mathcal{L}_{t 2 i}\right).
\end{equation}

The loss aims to achieve image-text alignment in the special domain for the discriminative task. 

\textbf{Text-conditioned Image Synthesis Loss.}  
Given the image feature $V_a$ and its corresponding token-level text feature $T_f$, we add $t$ steps’ random Gaussian noise into $V_a$, denoted as $V_{a,t}$ and incorporate text conditioning to denoise the $V_{a,t}$ by eliminating the $t$-step noise.  
Following equation~\ref{eq:is_gene}, the loss function of the image generation $\mathcal{L}_{\text {is}}$ to predict the noise at the $t$-time step is formulated as follows:  
\begin{equation}
\label{eq:is_gene_loss}
    \mathcal{L}_{\text {is}}=\mathbb{E}_{x_0, t, \epsilon}\left[\left\|\epsilon-\epsilon_\theta \left(V_{a, t}, t, T_t \right) \right\|^2\right].
\end{equation}


\textbf{Reciprocal Semantic Consistency Loss.} 
In a previous study on image manipulation~\cite{gal2022stylegan}, CLIP-guided diffusion was proposed as a method to guide the reverse diffusion process using a pre-trained CLIP model. This guidance was achieved through the utilization of the following loss function:

\begin{equation}
\label{eq:dclip}
\mathcal{L}_{DCLIP} = 1 - \frac{\Delta V \cdot \Delta T}{|\Delta V||\Delta T|},
\end{equation}

where $\Delta V = V_1 - V_2$ and $\Delta T = T_1 - T_2$ represent the differences between the source image $V_1$ and the target image $V_2$, and between the source text $T_1$ and the target text $T_2$, respectively.

The objective of the loss function \ref{eq:dclip} is to retain the maximum semantic information learned from the source data. 
Drawing inspiration from this loss, we introduce the feature $V_u$ with fused  semantic information outputted from the proposed fusion module, as well as the feature $V_a^{'}$ generated by feeding the $V_a$ of the visual autoencoder into the first linear layer of the fusion module. 
These features are utilized to enforce the reconstruction of the image feature $V_r$, which is predicted in a single step by calculating $\left(V_{a,t}-\sqrt{1-\bar{\alpha}_t} \epsilon_\theta\left(V_{a,t}, t, T_f\right)\right) / \sqrt{\bar{\alpha}_t}$, and $V_r$ has also projected into the feature space using the first linear layer of the fusion module. 
The optimization function is defined as follows:  
\begin{equation}
\label{eq:dclip1}
\mathcal{L}_{rsc} = 1 -  \frac{1}{n \times n_{t}} \sum_{i=1}^{n \times n_{t}} \frac{(V_{u_i} - V_{r_i}) \cdot (V_{a_i}^{'} - V_{r_i})}{|V_{u_i} - V_{r_i}||V_{a_i}^{'} - V_{r_i}|}, 
\end{equation}
where $n_t$ denotes the first dimension of $V_u$, $V_r$, and $V_a$. 
It's worth noting that since the first dimension of $V_u$ does not fully correspond to that of $V_r$ in the semantic, we first compute the average of $V_u$ along the first dimension and then repeat it $n_t$ times along the same dimension for consistency. 

This approach enables CLIP-enhanced semantic guidance by striking a balance between fused semantic information and token-level semantic information, as achieved through $\mathcal{L}_{rsc}$.

\section{Experiment}
\label{sec:experiment}

\subsection{Datasets}
In this paper, we focus on personalized design and creation, specifically within the fashion and product domains. 
To build our personalized training and testing datasets, we select two publicly available datasets.  
The CM-Fashion dataset \cite{zhang2022armani} comprises 500,000 images encompassing various garment categories (e.g., t-shirts, jackets, dresses) and includes detailed descriptions.   
From this dataset, we choose two subsets, namely Fashion-man and Fashion-woman, and create the training set with 100 samples per class and the testing set with 20 samples per class.   
Additionally, we utilize the M5Product dataset \cite{dong2022m5product}, which consists of over 5 million image-text pairs across more than 6,000 categories.   
For the M5Product dataset, we randomly sample 10 categories and construct the training and testing sets with an equal number of samples as in the fashion domain.   
For further information regarding the datasets, please refer to the supplementary materials. 

\subsection{Implement Details}
We construct our UniDiff model by utilizing Stable Diffusion v1-4~\cite{rombach2022high} and pre-trained CLIP-ViT-L/14~\cite{radford2021learning} as the base models.  
 The model is fine-tuned on three specific datasets, namely Fashion-man, Fashion-woman, and E-commerce Product datasets.  
The input resolutions of the visual autoencoder for generation and the visual encoder for discrimination are 512 $\times$ 512 and 224 $\times$ 224, respectively.  
We employ a batch size of 6 and fine-tune the models for 200 epochs.  
The AdamW~\cite{loshchilov2018decoupled} optimizer with a learning rate of 1$\textit{e}$-4 is employed.  
We utilize gradient accumulation of 8 steps in our training process. 
For $V_u$, we set the channel size to 4, matching the number of channels in SD's autoencoder.  
Additionally, we incorporate the exponential moving average (EMA) with a decay coefficient of 0.9999 at each iteration. The EMA model is used to evaluate the performance on downstream tasks. 
In the supplementary materials,  more experimental details are provided. 

\noindent \textbf{Evaluation.} We evaluate the performance of our UniDiff and compare it with existing methods on cross-modal retrieval and text-to-image generation tasks.  
For the retrieval tasks, we employ the Recall@$N$ metric to assess the performance. 
For text-to-image generation, we utilize the Fréchet Inception Distance (FID)~\cite{obukhov2020torchfidelity} metric to quantify the quality and diversity of the reference image bank. 
We use DDIM sampler~\cite{song2020denoising} with 50 sampling steps under the classifier-free guidance scale~\cite{ho2022classifier-free} of 5.

\subsection{Experimental Results} 
\label{subsec:experimental_rs}
\noindent\textbf{Discriminative Task.} Tables \ref{tab:main:man}, \ref{tab:main:woman}, and \ref{tab:main:product} display the experimental results of cross-modal retrieval tasks, including image-to-text retrieval and text-to-image retrieval, on Fashion-man, Fashion-woman, and E-commerce product datasets.   
Our proposed UniDiff consistently outperforms all CLIP models in these tasks, achieving higher accuracy.  
In particular, when fine-tuned with a small batch size, CLIP models suffer from modeling collapse, such as the drop from 60.52 to 35.79 in R@1 of image-to-text retrieval presented in Table~\ref{tab:main:man}. 
In contrast, UniDiff leverages the strong constraint of semantic consistency learning, achieving nearly a 10$\%$ performance gain. 


\begin{table*}[bt!]
\Large
\caption{Results of image-text retrieval on the Fashion-man dataset.}
\centering
\label{tab:main:man}
\vspace{-2mm}
\resizebox{0.9\columnwidth}{!}{
\begin{tabular}{cl | cccc | cccc}
\toprule[1pt]
 & \multirow{2}{*}{Model} & \multicolumn{4}{|c|}{Image-to-Text} &\multicolumn{4}{|c}{Text-to-Image}\\
 &      &{R@1} & {R@5} & {R@10} & {Mean R} & {R@1}  &{R@5} & {R@10}& {Mean R} \\
\midrule
 \multirow{3}{*}{Zero-Shot} &
 CLIP-ViT-B/32~\cite{radford2021learning}  &52.77  &84.50  &93.36 & 76.88 &45.76  &79.70  &90.04 & 71.83 \\
 & CLIP-ViT-B/16~\cite{radford2021learning}  &61.99    &88.93   &94.83  &81.92  &47.23  &78.59  &90.77  &72.20 \\
 & CLIP-ViT-L/14~\cite{radford2021learning}  &60.52   &90.03   &\textbf{97.05}  &82.53  &59.77  &85.98  &94.83 & 80.19 \\
\midrule
 \multirow{4}{*}{Finetuned} &
CLIP-ViT-B/32~\cite{radford2021learning}  & 27.31    &55.72   &67.53  &50.18  &28.00  &56.46  &69.37  &51.27 \\
&  CLIP-ViT-B/16~\cite{radford2021learning}  &32.47    &65.31   &75.65  &57.81  &32.84  &56.46  &67.53  &52.53\\
&  CLIP-ViT-L/14~\cite{radford2021learning} &35.79    &64.94   &77.86  &59.53  &37.64  &68.63  &78.59  &61.62 \\ 
& \textbf{UniDiff} &\textbf{70.48}    &\textbf{92.62}   &\textbf{97.05}  &\textbf{86.72}  &\textbf{65.68}  &\textbf{91.51}  &\textbf{97.42}  &\textbf{84.87} \\
\bottomrule[1pt]
\end{tabular}}
\vspace{-3mm}
\end{table*}

\begin{table*}[t!]
\Large
\caption{Results of image-text retrieval on the Fashion-woman dataset.}
\centering
\label{tab:main:woman}
\vspace{-2mm}
\resizebox{0.9\columnwidth}{!}{
\begin{tabular}{cl | cccc | cccc}
\toprule[1pt]
 &  \multirow{2}{*}{Model} & \multicolumn{4}{|c|}{Image-to-Text} &\multicolumn{4}{|c}{Text-to-Image}\\
 & & {R@1} & {R@5} & {R@10} & {Mean R} & {R@1}  &{R@5} & {R@10}& {Mean R} \\
\midrule
\multirow{3}{*}{Zero-Shot} & CLIP-ViT-B/32~\cite{radford2021learning}  &  61.70  &91.91   &96.59  &83.40  &57.02  &90.21  &94.46  &80.57 \\
 & CLIP-ViT-B/16~\cite{radford2021learning}  &65.11    &93.19   &95.74  &84.68  &57.87  & 91.91 &96.59  &82.13 \\
& CLIP-ViT-L/14~\cite{radford2021learning}  &70.21    &\textbf{96.17}   &\textbf{98.72}  &88.37  &71.06  &91.91 &\textbf{97.45}  &86.81 \\ 
\midrule
\multirow{4}{*}{Fine-tuned} &
CLIP-ViT-B/32~\cite{radford2021learning}  &45.53   &73.19   &83.83  &67.52  &43.83  &74.04 &84.26 &67.38  \\
& CLIP-ViT-B/16~\cite{radford2021learning}  &40.00    &72.77   &83.40  &65.39  &45.11  &74.89  &85.53  &68.51 \\
& CLIP-ViT-L/14~\cite{radford2021learning}  &50.21    &78.72   &88.51  &72.48  &47.23  &81.28 &88.51  &72.34 \\
& \textbf{UniDiff} & \textbf{73.62}   &95.32   & 98.29  &\textbf{89.08}  &\textbf{73.62}  &\textbf{94.04} &\textbf{97.45}  &\textbf{88.37} \\
\bottomrule[1pt]
\end{tabular}}
\vspace{-3mm}
\end{table*}

\begin{table*}[t!]
\Large
\caption{Results of image-text retrieval on the E-commerce product dataset.}
\centering
\label{tab:main:product}
\vspace{-3mm}
\resizebox{0.9\columnwidth}{!}{
\begin{tabular}{cl | cccc | cccc}
\toprule[1pt]
 & \multirow{2}{*}{Model} & \multicolumn{4}{|c|}{Image-to-Text} &\multicolumn{4}{|c}{Text-to-Image}\\
 &  &{R@1} & {R@5} & {R@10} & {Mean R} & {R@1}  &{R@5} & {R@10}& {Mean R} \\
\midrule
 \multirow{3}{*}{Zero-Shot}
& CLIP-ViT-B/32~\cite{radford2021learning}  &19.00    &46.00   &69.50  &44.80  &18.00  &52.50 &73.50  &48.00 \\ 
& CLIP-ViT-B/16~\cite{radford2021learning}  &19.50    &51.50   &73.50  &48.17  &19.00  &51.50 &75.00  &48.50 \\
& CLIP-ViT-L/14~\cite{radford2021learning}  &21.00    &55.00   &77.00  &51.00  &22.00  &56.50 &77.50  &53.00 \\
 \midrule
\multirow{4}{*}{Fine-tuned}
 & CLIP-ViT-B/32~\cite{radford2021learning}  &10.00    &23.50   &30.50  &21.33  &10.00  &24.50  &32.50  & 22.33\\
 & CLIP-ViT-B/16~\cite{radford2021learning}  &10.00    &22.00   &31.50  &21.16  &11.00  &21.00  &31.50  &21.16 \\
 & CLIP-ViT-L/14~\cite{radford2021learning}  & 19.00  &42.50   &51.00  &37.00  &22.00  &39.00  &51.00  &37.33 \\
 & \textbf{UniDiff} &\textbf{25.00}    &\textbf{61.00}   &\textbf{84.00}  &\textbf{56.67}  & \textbf{22.50} & \textbf{64.00}   & \textbf{85.50} &\textbf{57.33}\\
\bottomrule[1pt]
\end{tabular}}
\vspace{-4mm}
\end{table*}

\begin{table*}[t!]
\Large
\caption{Results of image synthesis on three datasets.}
\centering
\label{tab:main:generation}
\vspace{-2mm}
\resizebox{0.75\columnwidth}{!}{
\begin{tabular}{l |ccc}
\toprule[1pt]
 \multirow{2}{*}{Model} & \multicolumn{3}{|c}{FID} \\
 & Fashion-man & Fashion-woman & E-commerce Product\\
\midrule
Stable Diffusion~\cite{rombach2022high}  &25.40   &40.71   &14.03   \\ 
$\text{Stable Diffusion}_{\text{FT}}$~\cite{rombach2022high}  &16.32   &21.04   & 9.42   \\ 
Adapt-Parallel~\cite{chen2022adaptformer} &32.32   & 42.21  &25.55  \\
BitFit~\cite{zaken2021bitfit} &24.32   &30.73   &12.84  \\ 
LoRA-R4~\cite{hulora}  &23.03   &33.39   &14.23  \\
LoRA-R8~\cite{hulora}  &22.00   &34.21   &13.88  \\
\textbf{UniDiff} &\textbf{12.62}    &\textbf{15.97}   &\textbf{7.13}  \\
\bottomrule[1pt]
\end{tabular}}
\vspace{-3mm}
\end{table*}

\begin{table*}[t!]
\huge
\caption{Ablation study of loss functions on the Fashion-man dataset.}
\centering
\label{tab:aba:loss}
\vspace{-2mm}
\resizebox{0.7\columnwidth}{!}{
\begin{tabular}{l | ccc |ccc|c}
\toprule[1pt]
 \multirow{2}{*}{Model} & \multicolumn{3}{c|}{Image-to-Text} &\multicolumn{3}{c|}{Text-to-Image} &\multicolumn{1}{c}{Image Generation}\\
 &{R@1} & {R@5} & {R@10}  & {R@1}  &{R@5} & {R@10}& FID \\
\midrule
w$/$o ITC  & -   &-   &-  &-  &-  &- & 22.59 \\
w$/$o RSC  &60.51    &90.04   &\textbf{97.05}  &59.78  &85.98 &94.83  &25.44 \\ \midrule
\textbf{UniDiff}  &\textbf{70.48}   &\textbf{92.62}   &\textbf{97.05}   &\textbf{65.68}  &\textbf{91.51}  &\textbf{97.42} &\textbf{12.62} \\
\bottomrule[1pt]
\end{tabular}}
\vspace{-7mm}
\end{table*}

\begin{figure*}[t!]
\centering
\includegraphics[width=\linewidth]{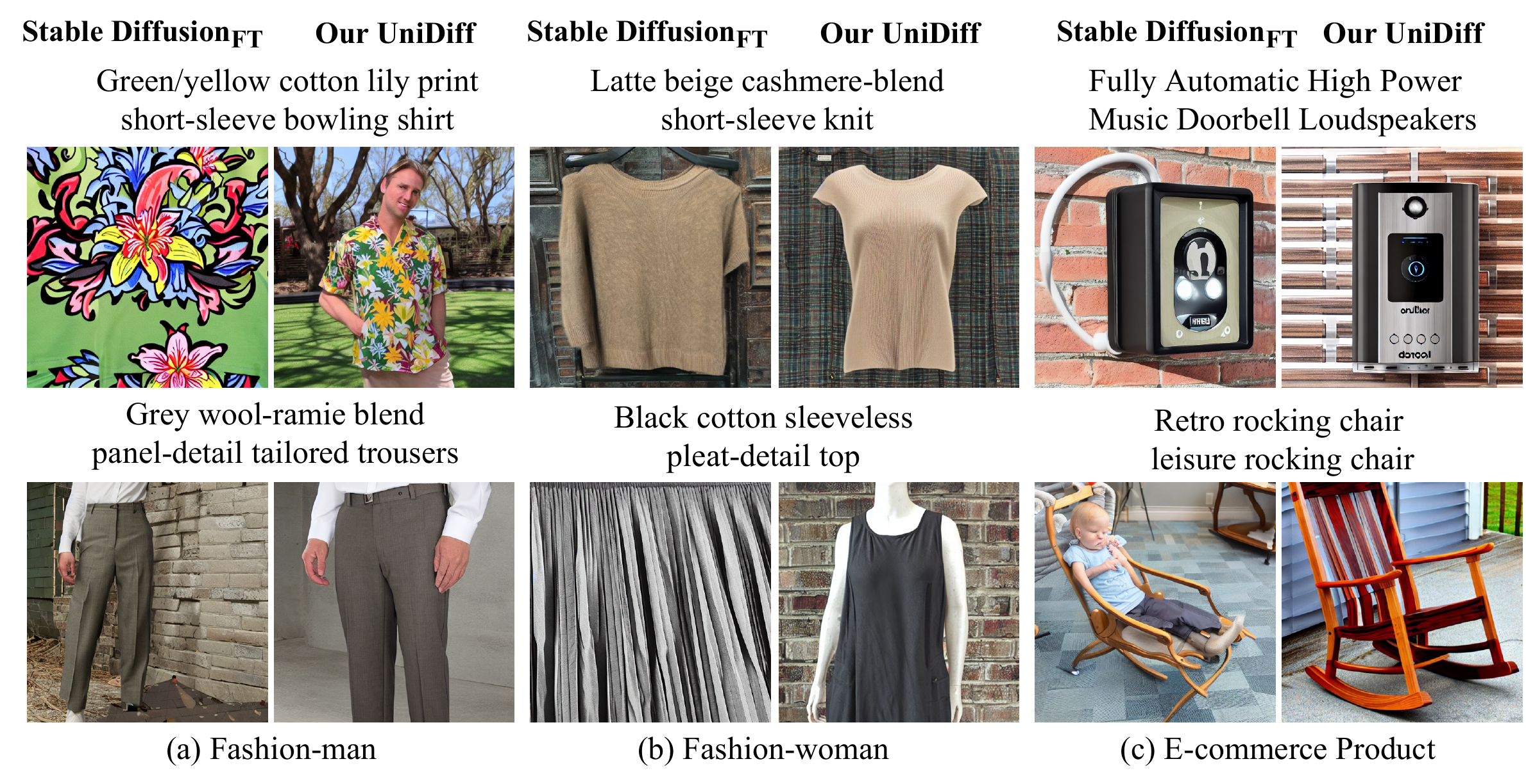}
  \vspace{-6mm}
  \caption{Qualitative comparisons of fine-tuned Stable Diffusion and our UniDiff on text-to-image generation of three datasets.}
  \vspace{-5mm}
\label{fig:aba:generated_image}
\end{figure*}

\noindent\textbf{Generative Task.} 
We conduct an experiment to compare the performance of Stable Diffusion (with and without fine-tuning) and three parameter-efficient fine-tuning methods (Adapt-Parallel, BitFit, and LoRA with the rank of 4~(LoRA-R4) or rank of 8~(LoRA-R8)) on text-to-image generation.  The results are summarized in Table~\ref{tab:main:generation}.  
Our proposed method, UniDiff, which incorporates reciprocal semantic consistency modeling, achieves the best FID scores on three distinct special domain datasets.  
We find that fully fine-tuning the Stable Diffusion is effective in adapting to the special domain data.  
However, LoRA performs poorly compared to the other methods, likely due to its inability to capture important descriptive details in text conditional image synthesis.  
 Figure~\ref{fig:aba:generated_image} shows the synthesized samples generated by the fine-tuned Stable Diffusion and UniDiff models, demonstrating that UniDiff's synthesized images exhibit higher realism and better text matching.  
Additional diverse generated images and visualized attention can be found in the supplementary materials, further supporting the superior performance of UniDiff. 


\begin{figure}[t!]
    \begin{subfigure}{0.48\textwidth}
        \centering
        \label{fig:man-guidance}
        \includegraphics[width=\textwidth]{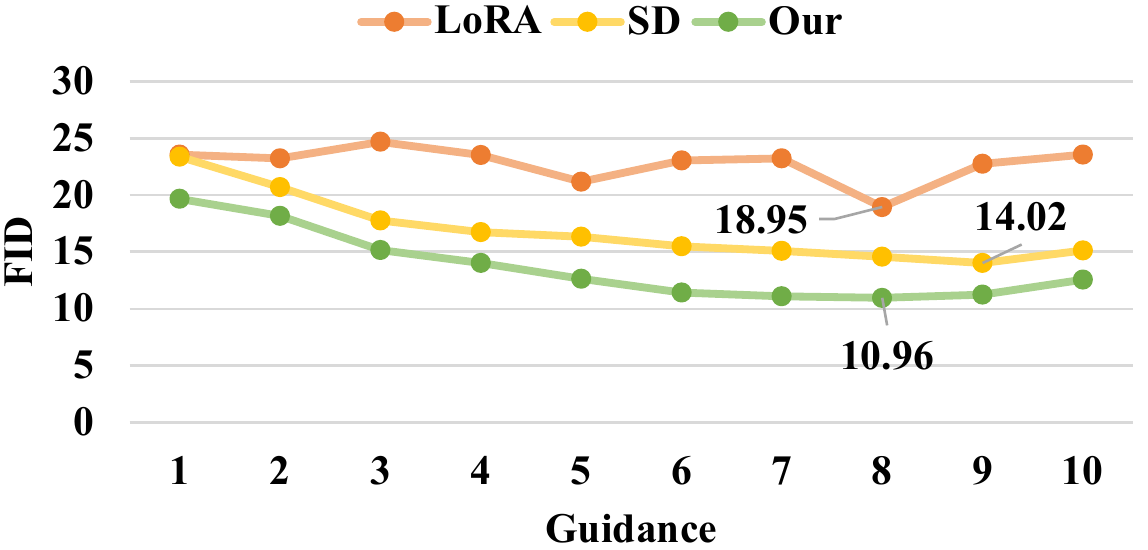}
        \caption{Fashion-man Dataset.}
    \end{subfigure}
    \begin{subfigure}{0.48\textwidth}
        \centering
        \label{fig:woman-guidance}
        \includegraphics[width=\textwidth]{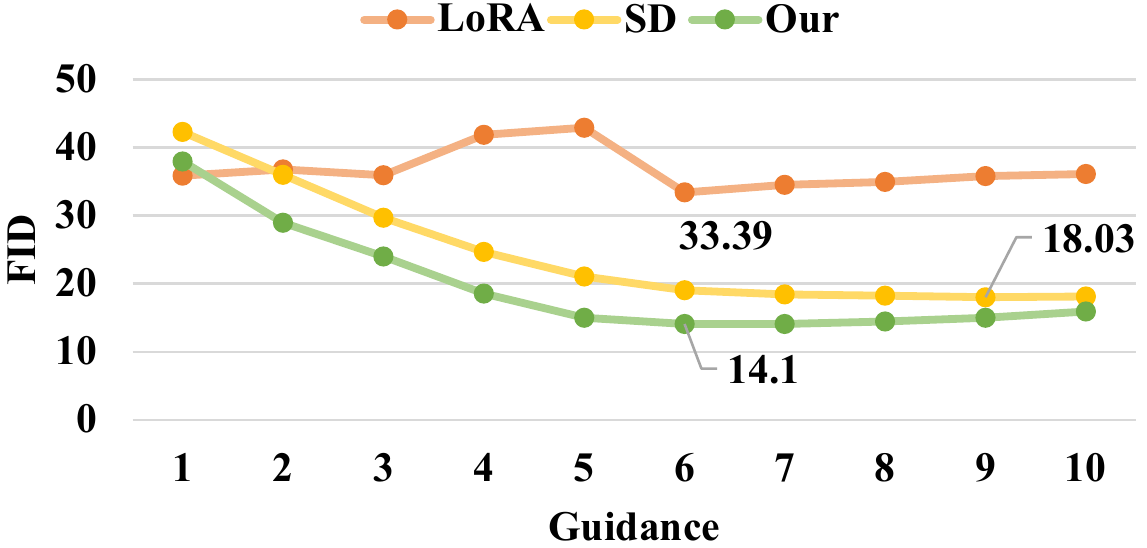}
        \caption{Fashion-woman Dataset.}
    \end{subfigure}
    \caption{Effect of different classifier-free guidance scales on different datasets.}
    \label{fig:aba:guidance}
    \vspace{-6mm}
\end{figure}


\subsection{Abaltion Study}  

\noindent \textbf{Loss Function.} We conduct ablation studies on the Fashion-man dataset to examine the impact of Image-Text Contrastive loss (ITC) and Reciprocal Semantic Consistency loss (RSC) on the performance of UniDiff in both discriminative and generative tasks.  
Table~\ref{tab:aba:loss} presents the results obtained when these loss components were excluded from the training process. 
The absence of the ITC loss leads to catastrophic results with a significant drop in performance.  
This emphasizes the importance of the ITC loss in providing aligned semantics in the new domain, facilitating effective cross-modal understanding. 
Similarly, excluding the RSC loss resulted in decreased performance.   
Its absence hindered the model's ability to retain important semantic information, leading to compromised performance. 
Our research studies demonstrate the significant contributions of both the ITC loss and the RSC loss in enhancing the discriminative and generative capabilities of UniDiff.   

\vspace{-1mm}
\noindent \textbf{Classifier-free Guidance Scales.} We examine the impact of different classifier-free guidance scales on text-conditioned image generation. 
Figure~\ref{fig:aba:guidance} illustrates that UniDiff outperforms fine-tuned SD and LoRA models, further evidencing the effectiveness of our reciprocal semantic consistency modeling.  

For more details of additional ablation experiments such as the impact of $V_u$ dimensions and different pooling methods, please refer to the supplementary materials. 
\vspace{-1mm}


\noindent \textbf{Limitation.} 
In the supplementary materials, we present a collection of failure cases and observe that certain textual descriptions result in unsatisfactory generated image quality.   
These limitations can be attributed to the constraints imposed by the provided prompts. 

\vspace{-3mm}
\section{Conclusion}
\label{sec:conclusion}
\vspace{-3mm}

In conclusion, this paper presents UniDiff, a novel unified multimodal model that integrates discriminative and generative capabilities in vision-language tasks. 
By incorporating reciprocal semantic consistency modeling, image-text contrastive learning, and text-conditioned image synthesis learning, UniDiff achieves significant improvements in vision-language retrieval and text-to-image generation.  
We highlight the importance of integrating discriminative and generative fine-tuning in vision and language, showcasing the synergistic effects between these capabilities.  
Through experiments on specific domain datasets, we validate UniDiff's effectiveness in personalized modeling and its broad distribution coverage.  
The results emphasize the potential of leveraging discriminative and generative capabilities in language and vision, opening new avenues for future research.  
Overall, this work contributes to the advancement of vision-language pre-training by exploring the integration of discriminative and generative capabilities.  
UniDiff establishes a robust pipeline for personalized modeling and serves as a benchmark for future studies in the field of vision-language interaction.
{
\small
\bibliographystyle{unsrt}
\bibliography{main_arxiv}

\begin{thebibliography}{10}

\bibitem{radford2021learning}
Alec Radford, Jong~Wook Kim, Chris Hallacy, Aditya Ramesh, Gabriel Goh,
  Sandhini Agarwal, Girish Sastry, Amanda Askell, Pamela Mishkin, Jack Clark,
  et~al.
\newblock Learning transferable visual models from natural language
  supervision.
\newblock In {\em International Conference on Machine Learning}, pages
  8748--8763. PMLR, 2021.

\bibitem{ddpm}
Jonathan Ho, Ajay Jain, and Pieter Abbeel.
\newblock Denoising diffusion probabilistic models.
\newblock {\em Advances in Neural Information Processing Systems},
  33:6840--6851, 2020.

\bibitem{adm}
Prafulla Dhariwal and Alexander Nichol.
\newblock Diffusion models beat gans on image synthesis.
\newblock {\em Advances in Neural Information Processing Systems},
  34:8780--8794, 2021.

\bibitem{rombach2022high}
Robin Rombach, Andreas Blattmann, Dominik Lorenz, Patrick Esser, and Bj{\"o}rn
  Ommer.
\newblock High-resolution image synthesis with latent diffusion models.
\newblock In {\em Proceedings of the IEEE/CVF Conference on Computer Vision and
  Pattern Recognition}, pages 10684--10695, 2022.

\bibitem{hulora}
Edward~J Hu, Yelong Shen, Phillip Wallis, Zeyuan Allen-Zhu, Yuanzhi Li, Shean
  Wang, Lu~Wang, and Weizhu Chen.
\newblock Lora: Low-rank adaptation of large language models.
\newblock {\em arXiv preprint arXiv:2106.09685}, 2021.

\bibitem{zaken2021bitfit}
Elad~Ben Zaken, Shauli Ravfogel, and Yoav Goldberg.
\newblock Bitfit: Simple parameter-efficient fine-tuning for transformer-based
  masked language-models.
\newblock {\em arXiv preprint arXiv:2106.10199}, 2021.

\bibitem{goodfellow2020generative}
Ian Goodfellow, Jean Pouget-Abadie, Mehdi Mirza, Bing Xu, David Warde-Farley,
  Sherjil Ozair, Aaron Courville, and Yoshua Bengio.
\newblock Generative adversarial networks.
\newblock {\em Communications of the ACM}, 63(11):139--144, 2020.

\bibitem{kim2022diffusionclip}
Gwanghyun Kim, Taesung Kwon, and Jong~Chul Ye.
\newblock Diffusionclip: Text-guided diffusion models for robust image
  manipulation.
\newblock In {\em Proceedings of the IEEE/CVF Conference on Computer Vision and
  Pattern Recognition}, pages 2426--2435, 2022.

\bibitem{kwon2022diffusion}
Gihyun Kwon and Jong~Chul Ye.
\newblock Diffusion-based image translation using disentangled style and
  content representation.
\newblock {\em arXiv preprint arXiv:2209.15264}, 2022.

\bibitem{chen2022adaptformer}
Shoufa Chen, Chongjian Ge, Zhan Tong, Jiangliu Wang, Yibing Song, Jue Wang, and
  Ping Luo.
\newblock Adaptformer: Adapting vision transformers for scalable visual
  recognition.
\newblock {\em arXiv preprint arXiv:2205.13535}, 2022.

\bibitem{radford2018improving}
Alec Radford, Karthik Narasimhan, Tim Salimans, Ilya Sutskever, et~al.
\newblock Improving language understanding by generative pre-training.
\newblock 2018.

\bibitem{devlin2019bert}
J.~Devlin, M.~Chang, K.~Lee, and K.~Toutanova.
\newblock {BERT: Pre-training of Deep Bidirectional Transformers for Language
  Understanding}.
\newblock In {\em North American Chapter of the Association for Computational
  Linguistics}, 2019.

\bibitem{liu2019roberta}
Y.~Liu, M.~Ott, N.~Goyal, J.~Du, M.~Joshi, D.~Chen, O.~Levy, M.~Lewis,
  L.~Zettlemoyer, and V.~Stoyanov.
\newblock {RoBERTa: A Robustly Optimized BERT Pretraining Approach}.
\newblock Preprint arXiv:1907.11692, 2019.

\bibitem{radford2019language}
Alec Radford, Jeffrey Wu, Rewon Child, David Luan, Dario Amodei, Ilya
  Sutskever, et~al.
\newblock Language models are unsupervised multitask learners.
\newblock {\em OpenAI blog}, 1(8):9, 2019.

\bibitem{jia2021scaling}
C.~Jia, Y.~Yang, Y.~Xia, Y.~Chen, Z.~Parekh, H.~Pham, Q.~Le, Y.~Sung, Z.~Li,
  and T.~Duerig.
\newblock Scaling up visual and vision-language representation learning with
  noisy text supervision.
\newblock In {\em International Conference on Machine Learning}, 2021.

\bibitem{yao2022filip}
L.~Yao, R.~Huang, L.~Hou, G.~Lu, M.~Niu, H.~Xu, X.~Liang, Z.~Li, X.~Jiang, and
  C.~Xu.
\newblock Filip: Fine-grained interactive language-image pre-training.
\newblock In {\em International Conference on Learning Representations}, 2022.

\bibitem{li2021supervision}
Y.~Li, F.~Liang, L.~Zhao, Y.~Cui, W.~Ouyang, J.~Shao, F.~Yu, and J.~Yan.
\newblock Supervision exists everywhere: A data efficient contrastive
  language-image pre-training paradigm.
\newblock Preprint arXiv:2110.05208, 2021.

\bibitem{gao2022pyramidclip}
Yuting Gao, Jinfeng Liu, Zihan Xu, Jun Zhang, Ke~Li, and Chunhua Shen.
\newblock Pyramidclip: Hierarchical feature alignment for vision-language model
  pretraining.
\newblock {\em arXiv preprint arXiv:2204.14095}, 2022.

\bibitem{lu2019vilbert}
Jiasen Lu, Dhruv Batra, Devi Parikh, and Stefan Lee.
\newblock Vilbert: Pretraining task-agnostic visiolinguistic representations
  for vision-and-language tasks.
\newblock {\em Advances in neural information processing systems}, 32, 2019.

\bibitem{tan2019lxmert}
Hao Tan and Mohit Bansal.
\newblock Lxmert: Learning cross-modality encoder representations from
  transformers.
\newblock In {\em Proceedings of the 2019 Conference on Empirical Methods in
  Natural Language Processing and the 9th International Joint Conference on
  Natural Language Processing (EMNLP-IJCNLP)}, pages 5100--5111, 2019.

\bibitem{chen2020uniter}
Yen-Chun Chen, Linjie Li, Licheng Yu, Ahmed El~Kholy, Faisal Ahmed, Zhe Gan,
  Yu~Cheng, and Jingjing Liu.
\newblock Uniter: Universal image-text representation learning.
\newblock In {\em European conference on computer vision}, pages 104--120.
  Springer, 2020.

\bibitem{lin2021m6}
Junyang Lin, Rui Men, An~Yang, Chang Zhou, Ming Ding, Yichang Zhang, Peng Wang,
  Ang Wang, Le~Jiang, Xianyan Jia, et~al.
\newblock M6: A chinese multimodal pretrainer.
\newblock Preprint arXiv:2103.00823, 2021.

\bibitem{ramesh2021zero}
Aditya Ramesh, Mikhail Pavlov, Gabriel Goh, Scott Gray, Chelsea Voss, Alec
  Radford, Mark Chen, and Ilya Sutskever.
\newblock Zero-shot text-to-image generation.
\newblock In {\em International Conference on Machine Learning}, pages
  8821--8831. PMLR, 2021.

\bibitem{wang2022beitv3}
Wenhui Wang, Hangbo Bao, Li~Dong, Johan Bjorck, Zhiliang Peng, Qiang Liu, Kriti
  Aggarwal, Owais~Khan Mohammed, Saksham Singhal, Subhojit Som, et~al.
\newblock Image as a foreign language: Beit pretraining for all vision and
  vision-language tasks.
\newblock {\em arXiv preprint arXiv:2208.10442}, 2022.

\bibitem{li2021albef}
Junnan Li, Ramprasaath Selvaraju, Akhilesh Gotmare, Shafiq Joty, Caiming Xiong,
  and Steven Chu~Hong Hoi.
\newblock Align before fuse: Vision and language representation learning with
  momentum distillation.
\newblock {\em Advances in Neural Information Processing Systems}, 34, 2021.

\bibitem{li2022blip}
Junnan Li, Dongxu Li, Caiming Xiong, and Steven Hoi.
\newblock Blip: Bootstrapping language-image pre-training for unified
  vision-language understanding and generation.
\newblock In {\em ICML}, 2022.

\bibitem{zhan2021product1m}
Xunlin Zhan, Yangxin Wu, Xiao Dong, Yunchao Wei, Minlong Lu, Yichi Zhang, Hang
  Xu, and Xiaodan Liang.
\newblock Product1m: Towards weakly supervised instance-level product retrieval
  via cross-modal pretraining.
\newblock In {\em Proceedings of the IEEE/CVF International Conference on
  Computer Vision}, pages 11782--11791, 2021.

\bibitem{dong2022m5product}
Xiao Dong, Xunlin Zhan, Yangxin Wu, Yunchao Wei, Michael~C Kampffmeyer,
  Xiaoyong Wei, Minlong Lu, Yaowei Wang, and Xiaodan Liang.
\newblock M5product: Self-harmonized contrastive learning for e-commercial
  multi-modal pretraining.
\newblock In {\em Proceedings of the IEEE/CVF Conference on Computer Vision and
  Pattern Recognition}, pages 21252--21262, 2022.

\bibitem{maaloe2019biva}
Lars Maal{\o}e, Marco Fraccaro, Valentin Li{\'e}vin, and Ole Winther.
\newblock Biva: A very deep hierarchy of latent variables for generative
  modeling.
\newblock {\em Advances in neural information processing systems}, 32, 2019.

\bibitem{razavi2019generating}
Ali Razavi, Aaron Van~den Oord, and Oriol Vinyals.
\newblock Generating diverse high-fidelity images with vq-vae-2.
\newblock {\em Advances in neural information processing systems}, 32, 2019.

\bibitem{ho2019flow++}
Jonathan Ho, Xi~Chen, Aravind Srinivas, Yan Duan, and Pieter Abbeel.
\newblock Flow++: Improving flow-based generative models with variational
  dequantization and architecture design.
\newblock In {\em International Conference on Machine Learning}, pages
  2722--2730. PMLR, 2019.

\bibitem{chen2020vflow}
Jianfei Chen, Cheng Lu, Biqi Chenli, Jun Zhu, and Tian Tian.
\newblock Vflow: More expressive generative flows with variational data
  augmentation.
\newblock In {\em International Conference on Machine Learning}, pages
  1660--1669. PMLR, 2020.

\bibitem{chen18PixelSNAIL}
XI~Chen, Nikhil Mishra, Mostafa Rohaninejad, and Pieter Abbeel.
\newblock {P}ixel{SNAIL}: An improved autoregressive generative model.
\newblock In Jennifer Dy and Andreas Krause, editors, {\em Proceedings of the
  35th International Conference on Machine Learning}, volume~80 of {\em
  Proceedings of Machine Learning Research}, pages 864--872. PMLR, 10--15 Jul
  2018.

\bibitem{parmar2018image}
Niki Parmar, Ashish Vaswani, Jakob Uszkoreit, Lukasz Kaiser, Noam Shazeer,
  Alexander Ku, and Dustin Tran.
\newblock Image transformer.
\newblock In {\em International conference on machine learning}, pages
  4055--4064. PMLR, 2018.

\bibitem{karras2020analyzing}
Tero Karras, Samuli Laine, Miika Aittala, Janne Hellsten, Jaakko Lehtinen, and
  Timo Aila.
\newblock Analyzing and improving the image quality of stylegan.
\newblock In {\em Proceedings of the IEEE/CVF conference on computer vision and
  pattern recognition}, pages 8110--8119, 2020.

\bibitem{biggan}
Andrew Brock, Jeff Donahue, and Karen Simonyan.
\newblock Large scale gan training for high fidelity natural image synthesis.
\newblock In {\em International Conference on Learning Representations}.

\bibitem{nichol2021glide}
Alex Nichol, Prafulla Dhariwal, Aditya Ramesh, Pranav Shyam, Pamela Mishkin,
  Bob McGrew, Ilya Sutskever, and Mark Chen.
\newblock Glide: Towards photorealistic image generation and editing with
  text-guided diffusion models.
\newblock {\em arXiv preprint arXiv:2112.10741}, 2021.

\bibitem{adapt1}
Runxin Xu, Fuli Luo, Zhiyuan Zhang, Chuanqi Tan, Baobao Chang, Songfang Huang,
  and Fei Huang.
\newblock Raise a child in large language model: Towards effective and
  generalizable fine-tuning.
\newblock In Marie{-}Francine Moens, Xuanjing Huang, Lucia Specia, and
  Scott~Wen{-}tau Yih, editors, {\em EMNLP}, pages 9514--9528. Association for
  Computational Linguistics, 2021.

\bibitem{adapt2}
Neil Houlsby, Andrei Giurgiu, Stanislaw Jastrzebski, Bruna Morrone, Quentin
  de~Laroussilhe, Andrea Gesmundo, Mona Attariyan, and Sylvain Gelly.
\newblock Parameter-efficient transfer learning for {NLP}.
\newblock In Kamalika Chaudhuri and Ruslan Salakhutdinov, editors, {\em ICML},
  volume~97 of {\em Proceedings of Machine Learning Research}, pages
  2790--2799. {PMLR}, 2019.

\bibitem{adapt3}
Menglin Jia, Luming Tang, Bor{-}Chun Chen, Claire Cardie, Serge~J. Belongie,
  Bharath Hariharan, and Ser{-}Nam Lim.
\newblock Visual prompt tuning.
\newblock In Shai Avidan, Gabriel~J. Brostow, Moustapha Ciss{\'{e}},
  Giovanni~Maria Farinella, and Tal Hassner, editors, {\em ECCV}, volume 13693
  of {\em Lecture Notes in Computer Science}, pages 709--727. Springer, 2022.

\bibitem{oord2018representation}
Aaron van~den Oord, Yazhe Li, and Oriol Vinyals.
\newblock Representation learning with contrastive predictive coding.
\newblock {\em arXiv preprint arXiv:1807.03748}, 2018.

\bibitem{gal2022stylegan}
Rinon Gal, Or~Patashnik, Haggai Maron, Amit~H Bermano, Gal Chechik, and Daniel
  Cohen-Or.
\newblock Stylegan-nada: Clip-guided domain adaptation of image generators.
\newblock {\em ACM Transactions on Graphics (TOG)}, 41(4):1--13, 2022.

\bibitem{zhang2022armani}
Xujie Zhang, Yu~Sha, Michael~C Kampffmeyer, Zhenyu Xie, Zequn Jie, Chengwen
  Huang, Jianqing Peng, and Xiaodan Liang.
\newblock Armani: Part-level garment-text alignment for unified cross-modal
  fashion design.
\newblock In {\em Proceedings of the 30th ACM International Conference on
  Multimedia}, pages 4525--4535, 2022.

\bibitem{loshchilov2018decoupled}
Ilya Loshchilov and Frank Hutter.
\newblock Decoupled weight decay regularization.
\newblock In {\em International Conference on Learning Representations}, 2018.

\bibitem{obukhov2020torchfidelity}
Anton Obukhov, Maximilian Seitzer, Po-Wei Wu, Semen Zhydenko, Jonathan Kyl, and
  Elvis Yu-Jing Lin.
\newblock High-fidelity performance metrics for generative models in pytorch,
  2020.
\newblock Version: 0.3.0, DOI: 10.5281/zenodo.4957738.

\bibitem{song2020denoising}
Jiaming Song, Chenlin Meng, and Stefano Ermon.
\newblock Denoising diffusion implicit models.
\newblock {\em arXiv preprint arXiv:2010.02502}, 2020.

\bibitem{ho2022classifier-free}
Jonathan Ho and Tim Salimans.
\newblock Classifier-free diffusion guidance.
\newblock {\em arXiv preprint arXiv:2207.12598}, 2022.

\bibitem{pytorch}
Adam Paszke, Sam Gross, Soumith Chintala, Gregory Chanan, Edward Yang, Zachary
  DeVito, Zeming Lin, Alban Desmaison, Luca Antiga, and Adam Lerer.
\newblock Automatic differentiation in pytorch.
\newblock In {\em NIPS Workshop}, 2017.

\bibitem{IS}
Tim Salimans, Ian~J. Goodfellow, Wojciech Zaremba, Vicki Cheung, Alec Radford,
  and Xi~Chen.
\newblock Improved techniques for training gans.
\newblock In Daniel~D. Lee, Masashi Sugiyama, Ulrike von Luxburg, Isabelle
  Guyon, and Roman Garnett, editors, {\em Advances in Neural Information
  Processing Systems 29: Annual Conference on Neural Information Processing
  Systems 2016, December 5-10, 2016, Barcelona, Spain}, pages 2226--2234, 2016.

\bibitem{kid}
Mikolaj Binkowski, Danica~J. Sutherland, Michael Arbel, and Arthur Gretton.
\newblock Demystifying {MMD} gans.
\newblock In {\em 6th International Conference on Learning Representations,
  {ICLR} 2018, Vancouver, BC, Canada, April 30 - May 3, 2018, Conference Track
  Proceedings}. OpenReview.net, 2018.

\end{thebibliography}
}

\newpage
\appendix

\section{Detailed Setting}
Our models are implemented in Pytorch~\cite{pytorch}.  
To speed up training, we use mixed precision training.  
All models are trained on 8 Nvidia V100  GPUs on our workstations. 

\begin{figure*}[!htb]
\centering
\includegraphics[width=1\linewidth]{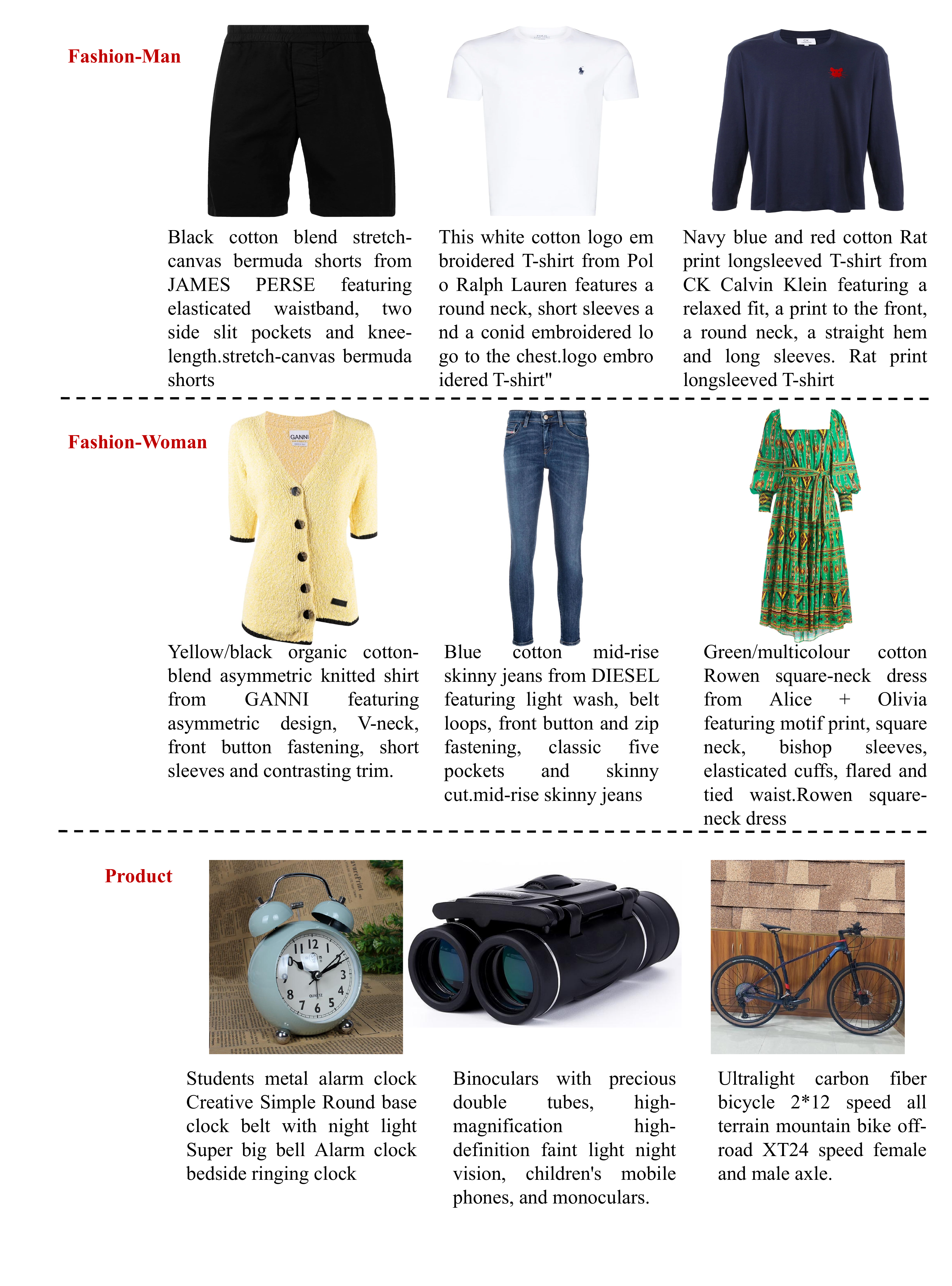}
  \caption{Some data samples on three datasets. }
\label{fig:appendix_demo}
\end{figure*}

\noindent \textbf{Dataset Details.} 
The CM-Fashion dataset comprises three subsets: Fashion-man, Fashion-woman, and Fashion-kid, encompassing more than 10 clothing categories. 
However, the Fashion-kid subset has a limited size, with only 465 samples available and some clothing categories represented by just two samples. 
Consequently, we concentrate our training efforts on the Fashion-man and Fashion-woman subsets. As previously mentioned in the dataset section, we select 100 samples from each category within these two subsets. 
The chosen categories for the Fashion-man subset include Shorts, Underwear and Socks, Sweaters and Knitwear, Sale Clothing, T-Shirts and Vests, Jackets, Jeans, Coats, Pants, Swimwear, Suits, Polo Shirts, Shirts, Sale Activewear, Athletic Jackets, and Activewear.  
Similarly, for the Fashion-woman subset, we select various categories such as Activewear, All-in-ones, Coats, Conscious activewear, Conscious clothing, Denim, Dresses, Footwear, Gym bags accessories, Jackets, Knitwear, Lingerie, Matching sets, Modest dressing, Pants, Sale activewear, Sale clothing, Shorts, Ski, Skirts, Suits tailoring, Swimwear, Swimwear beachwear, and Tops. 
Additionally, we also choose 10 categories from the M5Product dataset for the training and testing sets. 
These categories include Category Network Card, Mountain Bike, Leisure Chair, Gaming Console, Wireless Mouse, Timer, Air Conditioner, Bedside Cushion, Regular Telescope, and Smart Speaker.  
Figure~\ref{fig:appendix_demo} illustrates visual samples from our datasets.

\begin{table*}[!htb]
\Large
\caption{Results of image synthesis using Inception Score (IS) metric on three datasets.}
\centering
\label{tab:main:is}
\vspace{-2mm}
\resizebox{0.8\columnwidth}{!}{
\begin{tabular}{l |ccc}
\toprule[1pt]
 \multirow{2}{*}{Model} & \multicolumn{3}{|c}{Inception Score (IS) $\uparrow$} \\
 & Fashion-man & Fashion-woman & E-commerce Product\\
\midrule
Stable Diffusion~\cite{rombach2022high}  & 1.0064$\pm$0.0046  & 1.0065$\pm$0.0074   &1.0139$\pm$0.0054   \\ 
$\text{Stable Diffusion}_{\text{FT}}$~\cite{rombach2022high}  &\textbf{1.0687}$\pm$0.0180   & 1.0765$\pm$0.0311   & 1.0374$\pm$0.0212   \\ 
Adapt-Parallel~\cite{chen2022adaptformer}  &1.0000$\pm$0.0006   &1.0001$\pm$0.0001   &1.0000$\pm$0.0003  \\
BitFit~\cite{zaken2021bitfit} &1.0070$\pm$0.0037   & 1.0077$\pm$0.0090   &1.0141$\pm$0.0061 \\ 
LoRA-R4~\cite{hulora}  &1.0121$\pm$0.0099   & 1.0114$\pm$0.0107  &1.0117$\pm$0.0048  \\
LoRA-R8~\cite{hulora}  &1.0148$\pm$0.0076   & 1.0210$\pm$0.0124  &1.0218$\pm$0.0141  \\
\textbf{UniDiff} &1.0580$\pm$0.0146   &\textbf{1.0872}$\pm$0.0171  &\textbf{1.0420}$\pm$ 0.0121  \\
\bottomrule[1pt]
\end{tabular}}
\vspace{-3mm}
\end{table*}

\begin{table*}[!htb]
\Large
\caption{Results of image synthesis using Kernel Inception Distance (KID) metric on three datasets.}
\centering
\label{tab:main:kid}
\vspace{-2mm}
\resizebox{0.8\columnwidth}{!}{
\begin{tabular}{l |ccc}
\toprule[1pt]
 \multirow{2}{*}{Model} & \multicolumn{3}{|c}{Kernel Inception Distance (KID) $\downarrow$} \\
 & Fashion-man & Fashion-woman & E-commerce Product\\
\midrule
Stable Diffusion~\cite{rombach2022high}  &0.0155$\pm$0.0180  &0.0344$\pm$0.0283   &0.0071$\pm$0.0111   \\ 
$\text{Stable Diffusion}_{\text{FT}}$~\cite{rombach2022high}  &0.0058$\pm$0.0180   &  0.0121$\pm$0.0225 &0.0027$\pm$0.0105    \\ 
Adapt-Parallel~\cite{chen2022adaptformer} & 0.0153$\pm$0.0248    & 0.0309$\pm$0.0367    & 0.0137$\pm$0.0122   \\
BitFit~\cite{zaken2021bitfit} &0.0150$\pm$0.0233   &0.0340$\pm$0.0285   &0.0072$\pm$0.0111 \\ 
LoRA-R4~\cite{hulora}  &0.0139$\pm$0.0226   &0.0308$\pm$0.0300   &0.0086$\pm$0.0130  \\
LoRA-R8~\cite{hulora}  &0.0125$\pm$0.0229   &0.0270$\pm$0.0294   &0.0059$\pm$0.0105  \\
\textbf{UniDiff} &\textbf{0.0031}$\pm$0.0229 &\textbf{0.0041}$\pm$0.0171  &\textbf{0.0024}$\pm$0.0114  \\
\bottomrule[1pt]
\end{tabular}}
\vspace{-3mm}
\end{table*}

\begin{figure*}[!htb]
\centering
\includegraphics[width=1\linewidth]{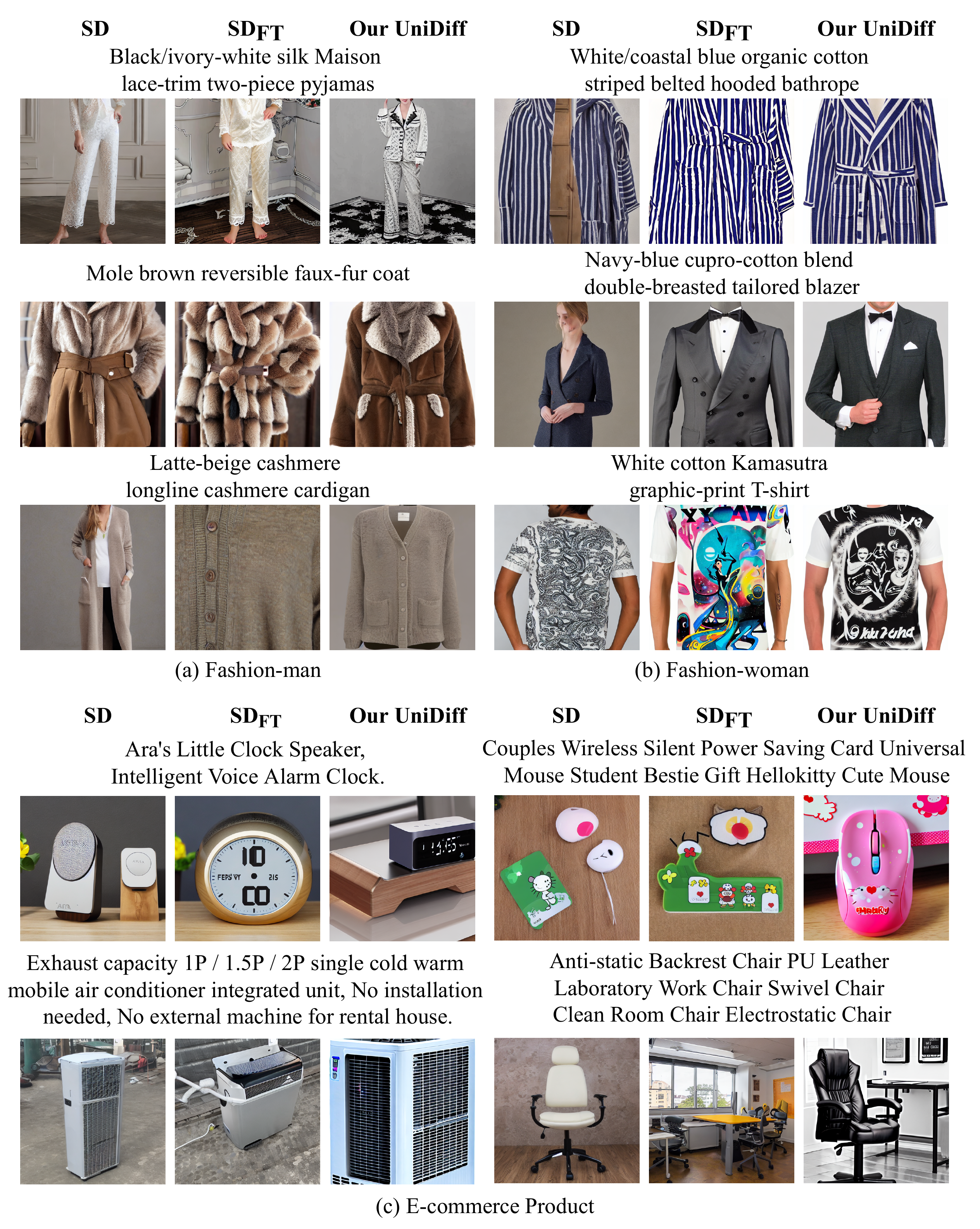}
  \caption{Further qualitative comparisons of fine-tuned Stable Diffusion and UniDiff on Text-to-Image Generation across Three Datasets. `SD' denotes the Stable Diffusion. }
\label{fig:more_syn}
\end{figure*}

\begin{figure*}[ht!]
\centering
\includegraphics[width=0.95\linewidth]{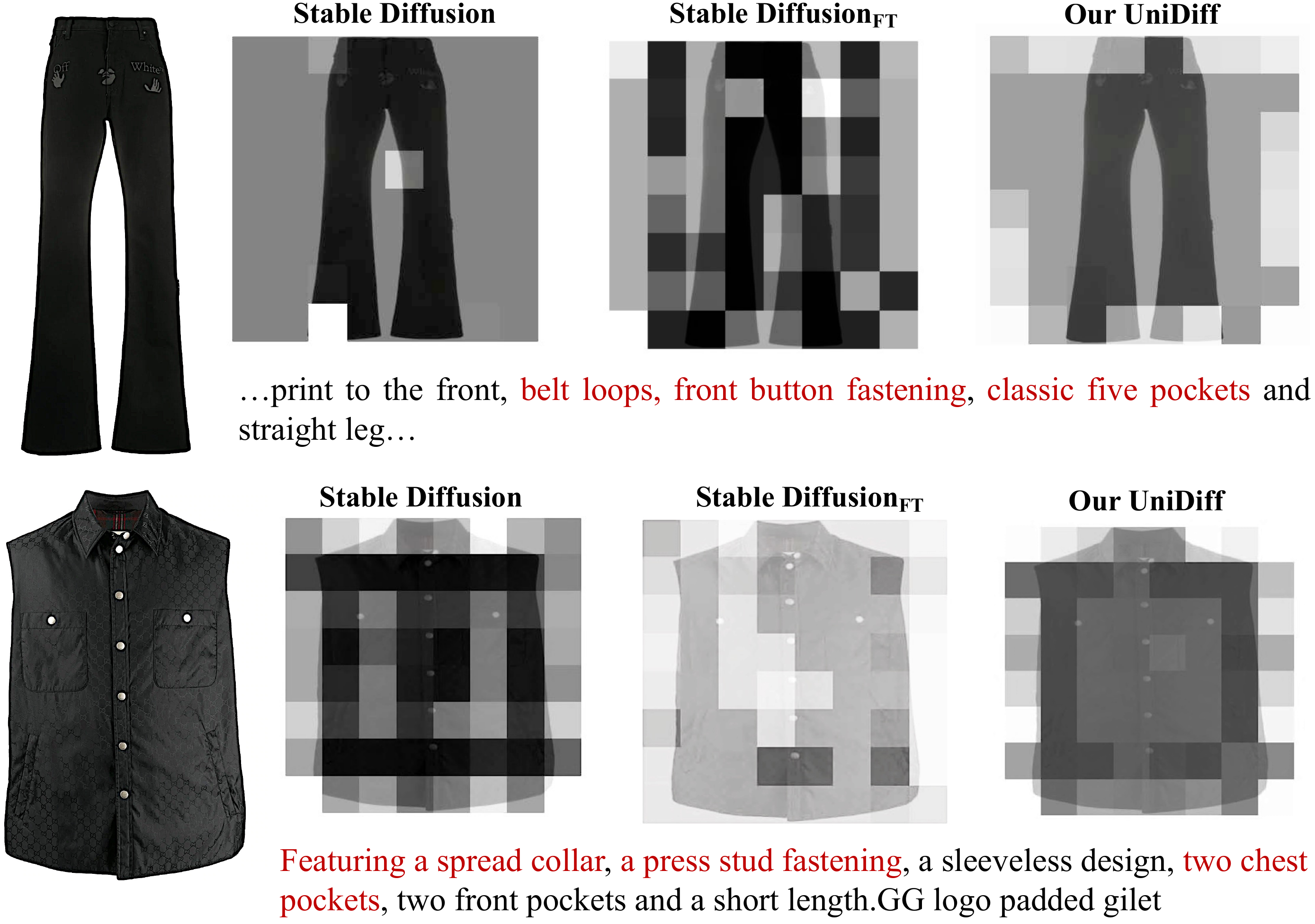}
  \caption{Attention attribution over cross-attention of U-NET learned by our UniDiff. }
\label{fig:atten}
\end{figure*}

\begin{table*}[!htb]
\huge
\caption{Impact of different dimensions of $V_u$ on the Fashion-man dataset in UniDiff. }
\centering
\label{tab:aba:dimen}
\resizebox{0.78\columnwidth}{!}{
\begin{tabular}{c | ccc |ccc|c}
\toprule[1pt]
 \multirow{2}{*}{Dimension} & \multicolumn{3}{c|}{Image-to-Text} &\multicolumn{3}{c|}{Text-to-Image} &\multicolumn{1}{c}{Image Generation}\\
 &{R@1} & {R@5} & {R@10}  & {R@1}  &{R@5} & {R@10}& FID \\
\midrule
4   &70.48   &92.62   &97.05   &65.68  &91.51  &97.42 &12.62  \\ \midrule
8  &67.89   &91.88  &97.05  &63.47  &91.88  &95.94 &14.36  \\
16 &68.26    &92.62  &97.04  &63.09  &91.89 &97.43  &12.40 \\ 
257 &63.47    &92.62  &96.68  &61.99  &89.29 &96.31  &11.47 \\ 
\bottomrule[1pt]
\end{tabular}}
\end{table*}

\begin{table*}[!htb]
\huge
\caption{Impact of different pooling approaches on the Fashion-man dataset in UniDiff.}
\centering
\label{tab:aba:pool}
\resizebox{0.86\columnwidth}{!}{
\begin{tabular}{l | ccc |ccc|c}
\toprule[1pt]
 \multirow{2}{*}{Approach} & \multicolumn{3}{c|}{Image-to-Text} &\multicolumn{3}{c|}{Text-to-Image} &\multicolumn{1}{c}{Image Generation}\\
 &{R@1} & {R@5} & {R@10}  & {R@1}  &{R@5} & {R@10}& FID \\
\midrule
AvgPooling  & 69.74  &93.73  &98.15  &64.58  &91.14  &96.68 &14.92  \\
AdaptiveAvgPooling  &71.59    &93.36  &98.15  &64.94  &90.04 &96.68  &14.19  \\
\midrule
MaxPooling &70.48   &92.62   &97.05   &65.68  &91.51  &97.42 &12.62 \\ 
\bottomrule[1pt]
\end{tabular}}
\end{table*}

\section{More Experimental Results using IS and KID Metrics.}
In addition to FID, we also utilize Inception Score (IS)\cite{IS} and Kernel Inception Distance (KID)\cite{kid} metrics to assess the performance of our model in the generation task, as presented in Table~\ref{tab:main:is} and Table~\ref{tab:main:kid}.  
The results from both tables indicate that our UniDiff model consistently achieves superior generative performance in terms of IS and KID metrics.  
This can be attributed to the effectiveness of our reciprocal semantic consistency modeling, further validating the capability of UniDiff to generate more realistic images.

\section{More Ablation Experiments.}
We investigate the influence of the $V_u$ dimension and various pooling methods, as outlined in Table~\ref{tab:aba:dimen} and Table~\ref{tab:aba:pool}. 
Regarding the dimension of $V_u$, we observe a pattern where the quality of image generation initially declines and then improves with increasing dimensionality.
However, in terms of retrieval performance, it consistently deteriorates with higher dimensions.
This can be attributed to the preservation of a substantial amount of semantic information at the image patch level as the dimension increases, thereby enhancing generation capabilities.
Nevertheless, some of this information exhibits redundancy, imposing limitations on retrieval tasks' improvement.  
Regarding the influence of different pooling methods, we observed that our model utilizing MaxPooling consistently achieves the best performance in both cross-modal retrieval and generation tasks. 
This can be attributed to the advantages of max pooling in capturing salient features within each pooling region.

\section{More Visualization}
We compared UniDiff with a fine-tuned stable diffusion model and a stable diffusion model without fine-tuning, as illustrated in Fig.~\ref{fig:more_syn}.  
Furthermore, the results clearly indicate that UniDiff's synthesized images exhibit higher realism and better text matching.   
To provide further evidence of the effectiveness of UniDiff, we present the attention map extracted from the cross-attention module of U-Net in Fig.~\ref{fig:atten}.   
Remarkably, the figure demonstrates that UniDiff, leveraging CLIP-feature and AutoEncoder features, effectively focuses on the key descriptions of the images.

\begin{figure*}[!htb]
\centering
\includegraphics[width=1\linewidth]{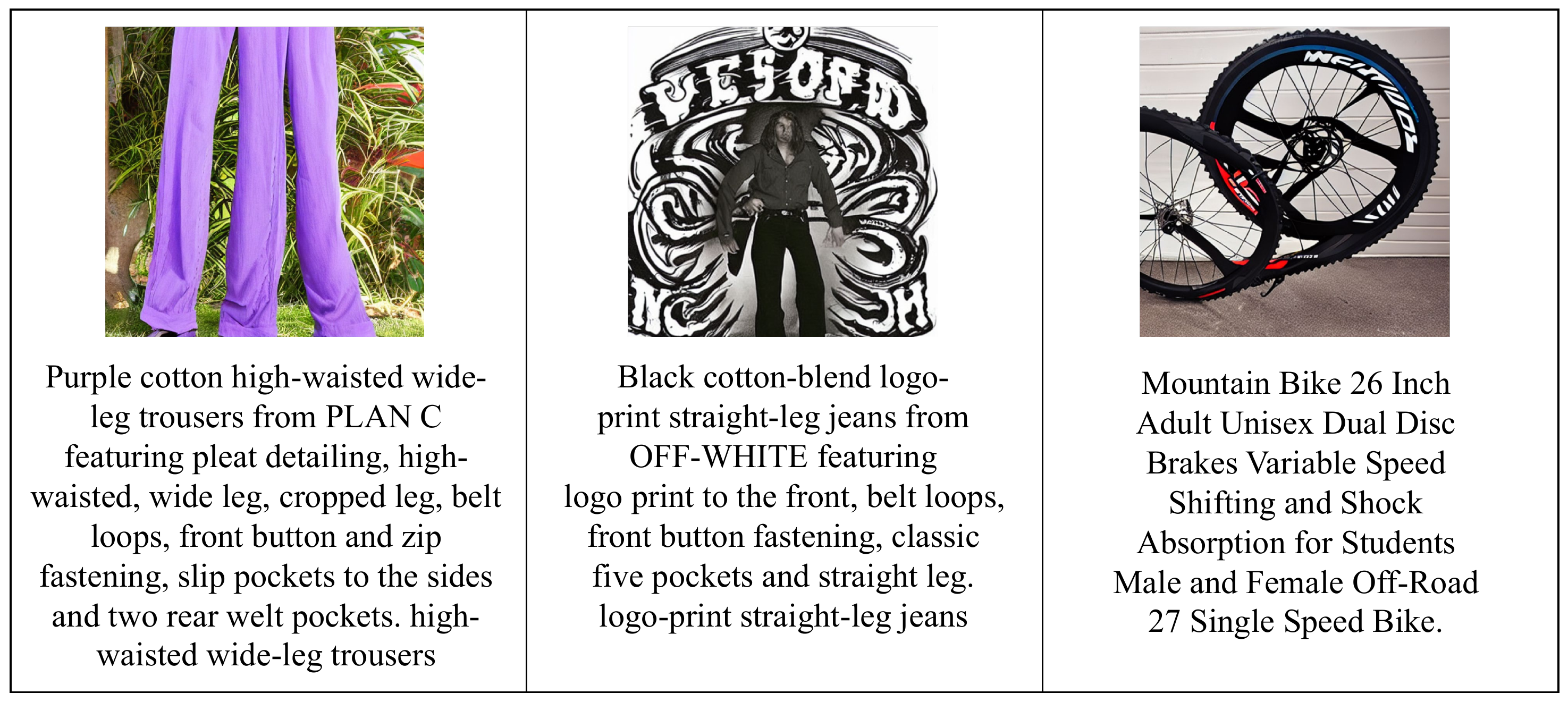}
  \caption{Failure cases in text-to-image generation. }
\label{fig:failure_case}
\end{figure*}

\begin{figure*}[!htb]
\centering
\includegraphics[width=1\linewidth]{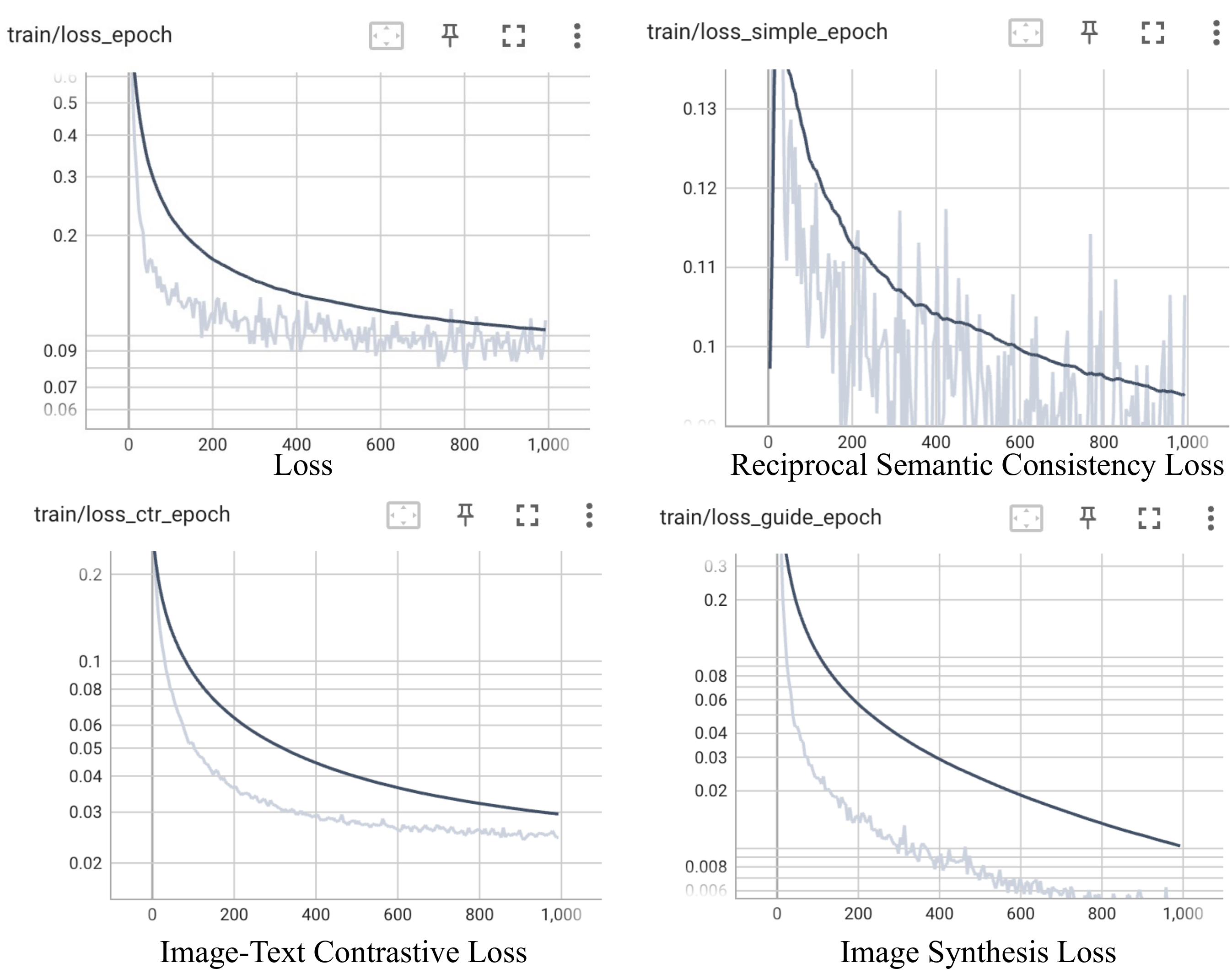}
  \caption{Loss function curves and individual loss components in UniDiff. }
\label{fig:our_loss}
\end{figure*}

\begin{figure*}[!htb]
\centering
\includegraphics[width=1\linewidth]{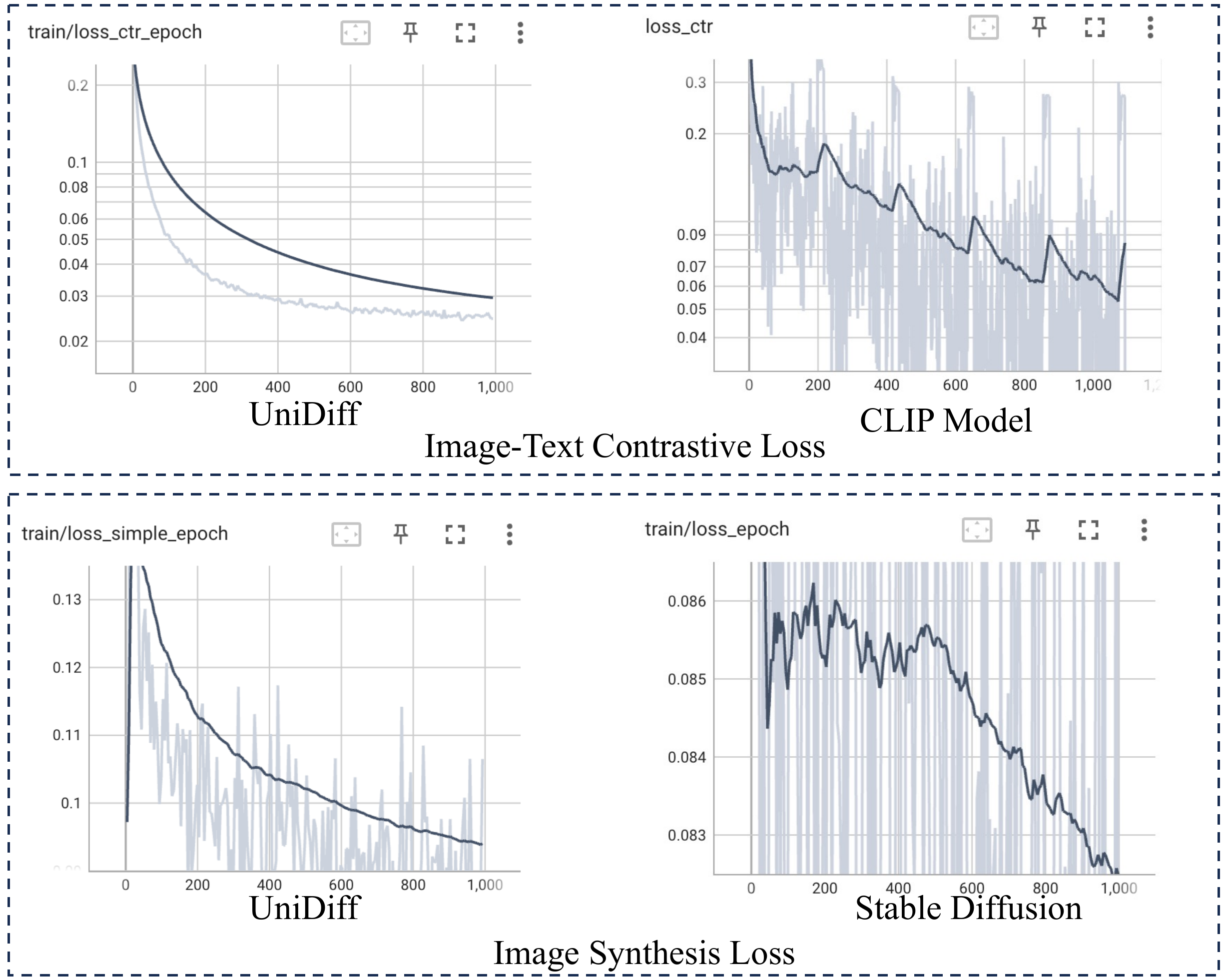}
  \caption{Comparison of loss function curves with CLIP model and Stable Diffusion model. }
\label{fig:loss_compare}
\end{figure*}

\section{Failure Case}
Figure~\ref{fig:failure_case} illustrates multiple instances of failure in UniDiff's generated outputs.  
These failures are predominantly attributed to the model's misinterpretation of textual prompts and the lack of detailed prompt descriptions.

\section{Loss Curve}
To assess the convergence of our model, we present Fig.~\ref{fig:our_loss}.  
Additionally, we evaluate the stability of our model by comparing it with separately fine-tuned ITC loss using CLIP and generation loss using stable diffusion, as depicted in Fig.~\ref{fig:loss_compare}.   
Notably, we observe that the curve of our UniDiff exhibits smoother and more stable compared to other methods. 

\end{document}